\documentclass{article}

\usepackage{times}
\usepackage{graphicx} % more modern
\usepackage{subfigure} 

\usepackage{natbib}

\usepackage{algorithm}
\usepackage{algorithmic}

\usepackage{hyperref}

\usepackage{amsmath}

\newcommand{\pd}[2]{\frac{\partial #1}{\partial #2}}

\newcommand{\mb}{\mathbf}
\newcommand{\mc}{\mathcal}
\newcommand{\argmin}{\operatornamewithlimits{argmin}}
\newcommand{\argmax}{\operatornamewithlimits{argmax}}
\newcommand{\median}{\operatornamewithlimits{median}}

\usepackage[accepted]{icml2014}

\icmltitlerunning{Sum of Functions Optimizer}

\begin{document} 

%\looseness=1
%\raggedbottom

\twocolumn[
\icmltitle{Fast large-scale optimization by unifying\\stochastic gradient and quasi-Newton methods}
%\icmltitle{Unifying stochastic gradient and quasi-Newton optimization methods in an adaptive low dimensional subspace}

% unless you've provided the [accepted] option to the icml2014
\icmlauthor{Jascha Sohl-Dickstein}{jascha@\{stanford.edu,khanacademy.org\}}
\icmlauthor{Ben Poole}{poole@cs.stanford.edu}
\icmlauthor{Surya Ganguli}{sganguli@stanford.edu}
%\icmladdress{Stanford University, Stanford, CA USA}

\icmlkeywords{optimization, quasi-Newton, stochastic gradient descent}

\vskip 0.3in
]

\begin{abstract} 
We present an algorithm for minimizing a sum of functions that combines the computational efficiency of stochastic gradient descent (SGD) with the second order curvature information leveraged by quasi-Newton methods. We unify these approaches by maintaining an independent Hessian approximation for each contributing function in the sum. We maintain computational tractability and limit memory requirements even for high dimensional optimization problems by storing and manipulating these quadratic approximations in a shared, time evolving, low dimensional subspace. Each update step requires only a single contributing function or minibatch evaluation (as in SGD), and each step is scaled using an approximate inverse Hessian and little to no adjustment of hyperparameters is required (as is typical for quasi-Newton methods). This algorithm contrasts with earlier stochastic second order techniques that treat the Hessian of each contributing function as a noisy approximation to the full Hessian, rather than as a target for direct estimation. We experimentally demonstrate improved convergence on seven diverse optimization problems. The algorithm is released as open source Python and MATLAB packages.
%TODO cite bengio ICLR submission?
%TODO motivate and discuss in terms of minibatches rather than subfunctions
\end{abstract}
\section{Introduction}
A common problem in optimization is to find a vector $\mb x^* \in \mc R^M$ which minimizes a function $F\left(\mb x\right)$, where $F\left(\mb x\right)$ is a sum of $N$ computationally cheaper differentiable subfunctions $f_i\left(\mb x\right)$,
\begin{align}
\label{eq F}
F\left(\mb x\right) &= \sum_{i=1}^N f_i\left(\mb x\right), \\
\label{eq Fmin}
\mb x^* &= \argmin_{\mb x} F\left(\mb x\right)
.
\end{align}
Many optimization tasks fit this form \cite{boyd2004convex}, including training of autoencoders, support vector machines, and logistic regression algorithms, as well as parameter estimation in probabilistic models. % \cite{MPF_ICML}.
In these cases each subfunction corresponds to evaluating the objective on a separate data minibatch, thus the number of subfunctions $N$ would be the datasize $D$ divided by the minibatch size $S$.  
This scenario is commonly referred to in statistics as M-estimation \cite{Huber81}.
%In statistics, $\mb x^*$ as defined by Equations \ref{eq F} and \ref{eq Fmin} is referred to as an M-estimator \cite{Huber81}.

\begin{figure*}[htp]
%\begin{center}
\centering
\begin{tabular}{ccc}
\hspace{-0.07\linewidth}
\begin{tabular}{c}
\includegraphics[width=0.35\linewidth]{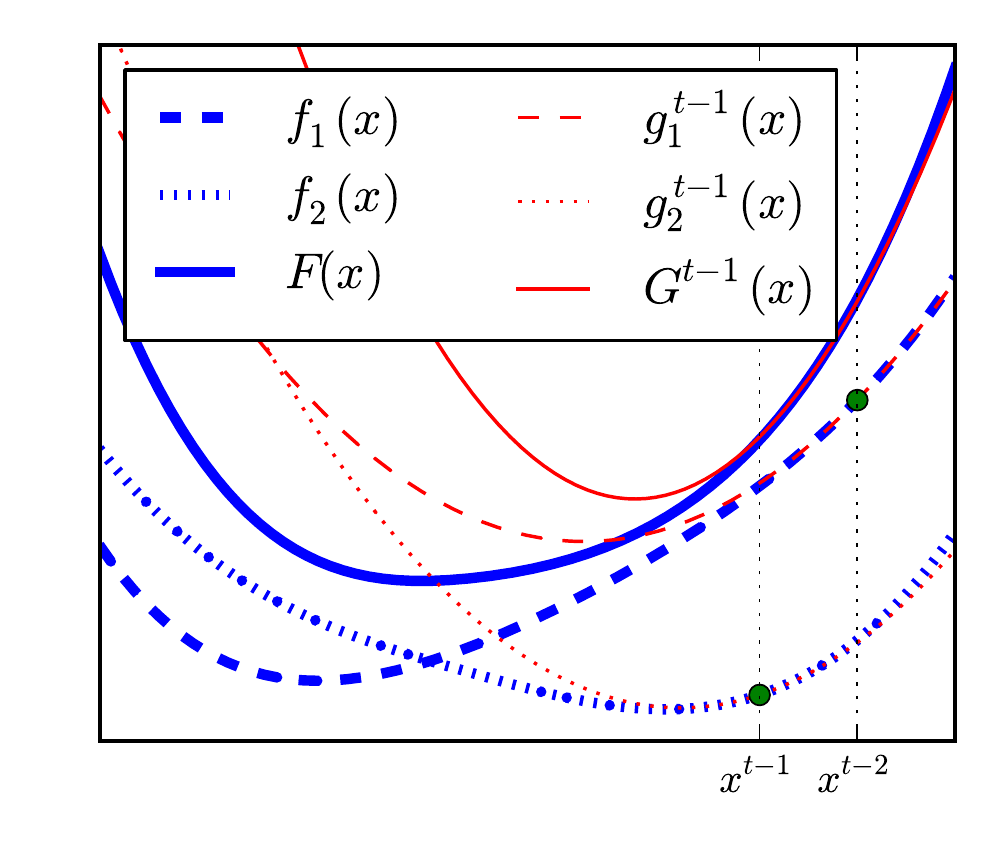} \vspace{-2mm} \\
(a)
\end{tabular}
 & 
\hspace{-0.04\linewidth}
\begin{tabular}{c}
\includegraphics[width=0.35\linewidth]{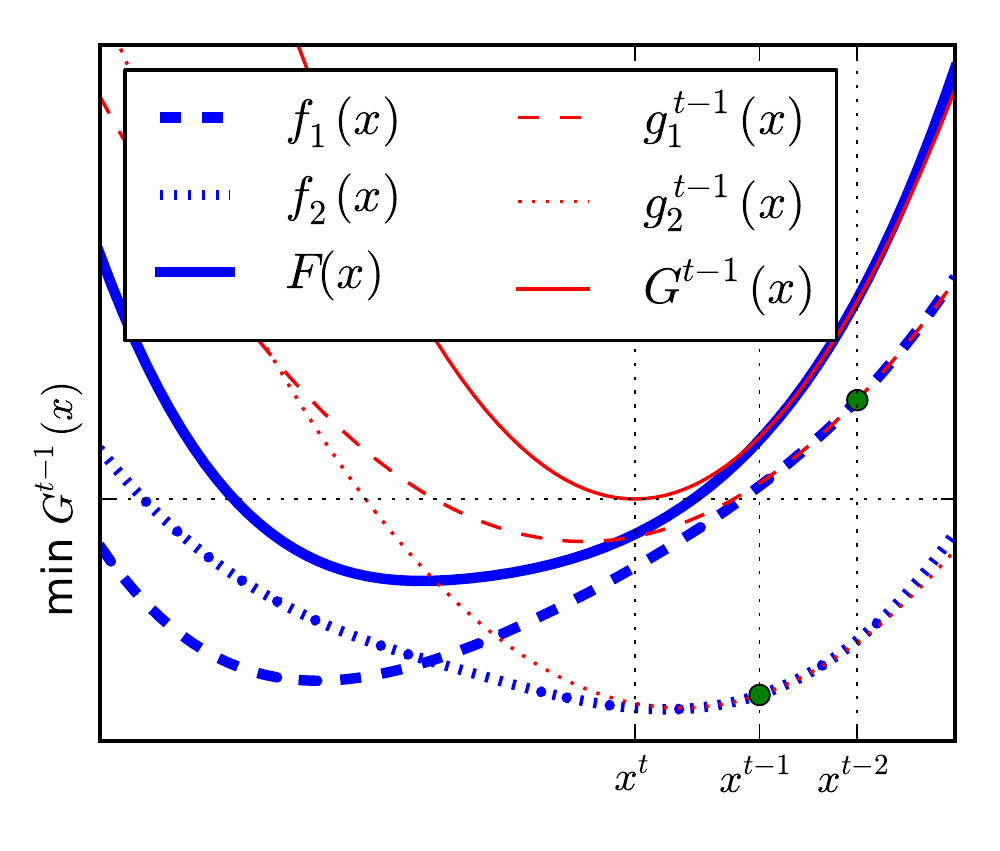} \vspace{-2mm} \\
(b)
\end{tabular}
 & 
\hspace{-0.06\linewidth}
\begin{tabular}{c}
\includegraphics[width=0.35\linewidth]{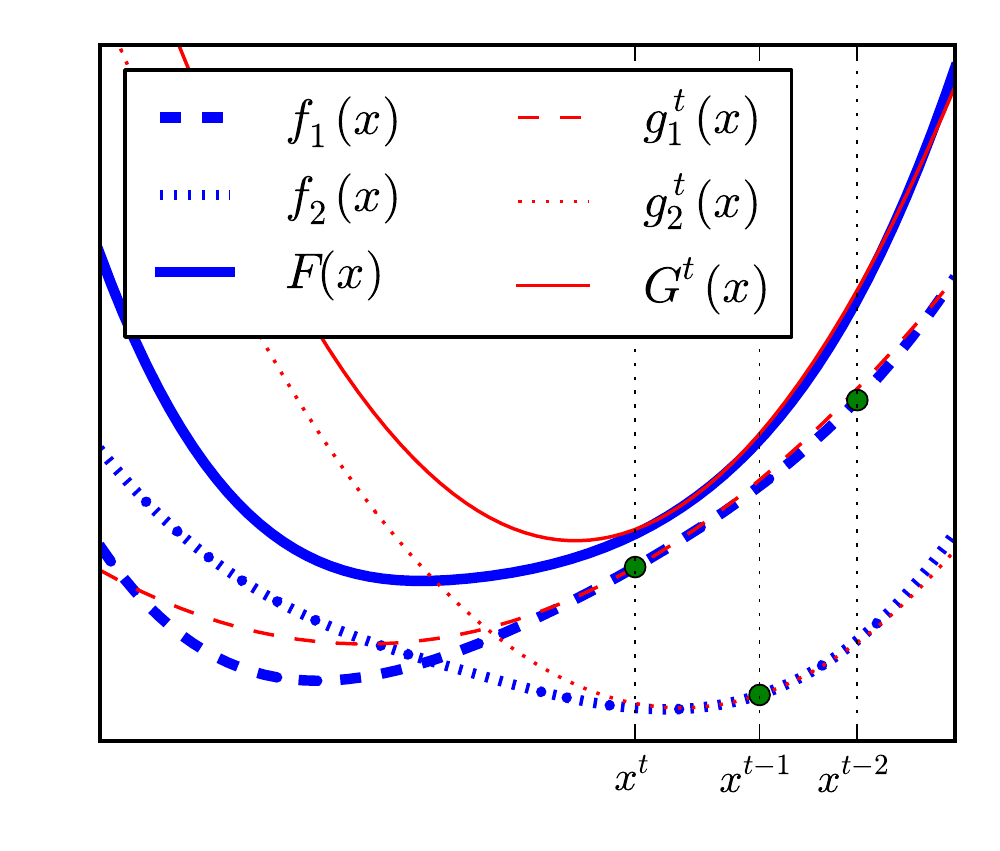} \vspace{-2mm} \\
(c)
\end{tabular}
\end{tabular}
%\end{center}
%\vspace{-2.5mm}
\caption{
A cartoon illustrating the proposed optimization technique.  {\em (a)} The objective function $F\left( x\right)$ (solid blue line) consists of a sum of two subfunctions (dashed blue lines), $F\left( x\right) = f_1\left(x\right) + f_2\left(x\right)$.  At learning step $t-1$, $f_1\left(x\right)$ and $f_2\left(x\right)$ are approximated by quadratic functions $g^{t-1}_1\left(x\right)$ and $g^{t-1}_2\left(x\right)$ (red dashed lines).  The sum of the approximating functions $G^{t-1}\left( x\right)$ (solid red line) approximates the full objective $F\left( x\right)$.  The green dots indicate the parameter values at which each subfunction has been evaluated  {\em (b)} The next parameter setting $x^t$ is chosen by minimizing the approximating function $G^{t-1}\left( x\right)$ from the prior update step.  See Equation \ref{eq x^t}.  {\em (c)} After each parameter update, the quadratic approximation for one of the subfunctions is updated using a second order expansion around the new parameter vector $x^t$.  See Equation \ref{eq subf upd}.  The constant and first order term in the expansion are evaluated exactly, and the second order term is estimated by performing BFGS on the subfunction's history.  In this case the approximating subfunction $g^{t}_1\left(x\right)$ is updated (long-dashed red line).  This update is also reflected by a change in the full approximating function $G^{t}\left( x\right)$ (solid red line).  %The function value $f_1\left(x\right)$ and gradient $f'_1\left(x\right)$ are computed exactly at $x^t$, and used in $g^{t}_1\left(x\right)$, while the Hessian is approximated via BFGS from the evaluation history of $f'_1\left(x\right)$. 
Optimization proceeds by repeating these two illustrated update steps.  In order to remain tractable in memory and computational overhead, optimization is performed in an adaptive low dimensional subspace determined by the history of gradients and iterates.
}
\label{fig cartoon}
\end{figure*}
There are two general approaches to efficiently optimizing a function of this form.  The first is to use a quasi-Newton method \cite{Dennis1977}, of which BFGS \cite{Broyden1970,Fletcher1970,Goldfarb1970,Shanno1970} or LBFGS \cite{Liu1989} are the most common choices.  Quasi-Newton methods use the history of gradient evaluations to build up an approximation to the inverse Hessian of the objective function $F\left(\mb x\right)$.  By making descent steps which are scaled by the approximate inverse Hessian, and which are therefore longer in directions of shallow curvature and shorter in directions of steep curvature, quasi-Newton methods can be orders of magnitude faster than steepest descent.  Additionally, quasi-Newton techniques typically require adjusting few or no hyperparameters, because they use the measured curvature of the objective function to set step lengths and directions.  However, direct application of quasi-Newton methods requires calculating the gradient of the full objective function $F\left(\mb x\right)$ at every proposed parameter setting $\mb x$, which can be very computationally expensive.%, especially for large $N$. 

The second approach is to use a variant of Stochastic Gradient Descent (SGD) \cite{Robbins1951,Bottou}.  In SGD, only one subfunction's gradient is evaluated per update step, and a small step is taken in the negative gradient direction.  More recent descent techniques like IAG \cite{Blatt2007}, SAG \cite{Roux2012}, and MISO \cite{Mairal2013,Mairal2014} instead take update steps in the average gradient direction.  For each update step, they evaluate the gradient of one subfunction, and update the average gradient using its new value.  \cite{Bach2013} averages the iterates rather than the gradients.  If the subfunctions are similar, then SGD can also be orders of magnitude faster than steepest descent on the full batch.  However, because a different subfunction is evaluated for each update step, the gradients for each update step cannot be combined in a straightforward way to estimate the inverse Hessian of the full objective function.  Additionally, efficient optimization with SGD typically involves tuning a number of hyperparameters, which can be a painstaking 
and frustrating process.  \cite{Ngiam2011} compares the performance of stochastic gradient and quasi-Newton methods on neural network training, and finds both to be competitive.

Combining quasi-Newton and stochastic gradient methods could improve optimization time, and reduce the need to tweak optimization hyperparameters.  %\cite{} use a trust region approach to .
This problem has been approached from a number of directions.  
In \cite{Schraudolph2007,Sunehag2009} a stochastic variant of LBFGS is proposed.  In \cite{Martens2010}, \cite{Byrd2011}, and \cite{Vinyals2011} stochastic versions of Hessian-free optimization are implemented and applied to optimization of deep networks. 
In \cite{Lin2008} a trust region Newton method is used to train logistic regression and linear SVMs using minibatches.  
In \cite{Hennig2013} a nonparametric quasi-Newton algorithm is proposed based on noisy gradient observations and a Gaussian process prior.  
In \cite{Byrd2014} LBFGS is performed, but with the contributing changes in gradient and position replaced by exactly computed Hessian vector products computed periodically on extra large minibatches. 
%implement versions of Hessian-free optimization where Hessian vector products are computed on only a subset of the data.  
Stochastic meta-descent \cite{Schraudolph1999}, AdaGrad \cite{Duchi2010}, and SGD-QN \cite{Bordes2009} rescale the gradient independently for each dimension, and can be viewed as accumulating something similar to a diagonal approximation to the Hessian.  %However, they are blind to off-diagonal terms in the Hessian.  
All of these techniques treat the Hessian on a subset of the data as a noisy approximation to the full Hessian.  To reduce noise in the Hessian approximation, they rely on regularization and very large minibatches to descend $F\left(\mb x\right)$.  % despite these noisy Hessian observations. 
Thus, unfortunately each update step requires the evaluation of many subfunctions and/or yields a highly regularized (i.e. diagonal) approximation to the full Hessian.    

%We will treat the Hessian for each subfunction as a direct target for estimation, rather than as a noisy approximation of another matrix.

% For example, both stochastic meta-descent \cite{Schraudolph1999} and AdaGrad \cite{Duchi2010} rescale the gradient independently for each dimension, and can be viewed as accumulating something similar to a diagonal approximation to the Hessian.  However, they are blind to off-diagonal terms in the Hessian. 
%Sophisticated regularization, backtracking, and line searches enable descent on the full objective despite working with subsets of the data.  However, all of these techniques require the use of a large enough subset that the Hessian vector product is a good approximation to the Hessian vector product for the full dataset.  
%Some algorithms designed for a single optimization problem take advantage of a decomposition of the objective function into a sum of subfunctions, for instance for support vector machines \cite{Bordes2009}.  
%Finally, although not strictly a quasi-Newton technique, the natural gradient is a powerful tool for incorporating incorporating higher order information into stochastic gradient update steps \cite{Amari1998,Pascanu2013}.

We develop a novel second-order quasi-Newton technique that only requires the evaluation of {\it a single} subfunction per update step. In order to achieve this substantial simplification, we treat the full Hessian of each subfunction as a direct target for estimation, thereby maintaining a separate quadratic approximation of each subfunction.  This approach differs from all previous work, which in contrast treats the Hessian of each subfunction as a noisy approximation to the full Hessian.  Our approach allows us to combine Hessian information from multiple subfunctions in a much more natural and efficient way than previous work, and avoids the requirement of large minibatches per update step to accurately estimate the full Hessian.  Moreover, we develop a novel method to maintain computational tractability and limit the memory requirements of this quasi-Newton method in the face of high dimensional optimization problems (large $M$).  We do this by storing and manipulating the subfunctions in a shared, adaptive low dimensional subspace, determined by the recent history of the gradients and iterates.

Thus our optimization method can usefully estimate and utilize powerful second-order information %inherent in the total function $F\left(\mb x\right)$ 
while simultaneously combatting two potential sources of computational intractability: large numbers of subfunctions (large N) and a high-dimensional optimization domain (large M).   Moreover,  the use of a second order approximation means that minimal or no adjustment of hyperparameters is required.  We refer to the resulting algorithm as Sum of Functions Optimizer (SFO).
%We additionally introduce a novel method of applying the natural gradient by performing a change of parameters, and a method to dynamically rescale the Hessian in order to decorrelate update steps. 
We demonstrate that the combination of techniques and new ideas inherent in SFO results in faster optimization on seven disparate example problems.  Finally, we release the optimizer and the test suite as open source Python and MATLAB packages.

\section{Algorithm}
Our goal is to combine the benefits of stochastic and quasi-Newton optimization techniques.  We first describe the general procedure by which we optimize the parameters $\mb x$.  We then describe the construction of the shared low dimensional subspace which makes the algorithm tractable in terms of computational overhead and memory for large problems. 
This is followed by a description of the BFGS method by which an online Hessian approximation is maintained for each subfunction.  Finally, we end this section with a review of implementation details.

\subsection{Approximating Functions}\label{sec approx}
We define a series of functions $G^t\left(\mb x\right)$ intended to approximate $F\left(\mb x\right)$,
\begin{align}
G^t\left(\mb x\right) = \sum_{i=1}^N g^t_i\left(\mb x\right)
\label{eq approximating functions}
,
\end{align}
where the superscript $t$ indicates the learning iteration.  Each $g^t_i\left(\mb x\right)$ serves as a quadratic approximation to the corresponding $f_i\left(\mb x\right)$.  The functions $g^t_i\left(\mb x\right)$ will be stored, and one of them will be updated per learning step.

\subsection{Update Steps}\label{sec update}

As is illustrated in Figure \ref{fig cartoon}, optimization is performed by repeating the steps:
\begin{enumerate}
  \item \label{minG} Choose a vector $\mb x^t$ by minimizing the approximating objective function $G^{t-1}\left(\mb x\right)$,
\begin{align}
\mb x^t = \argmin_{\mb x} G^{t-1}\left(\mb x\right)
.
\label{eq x^t}
\end{align}
Since $G^{t-1}\left(\mb x\right)$ is a sum of quadratic functions $g_i^{t-1}\left( \mb x \right)$, it can be exactly minimized by a Newton step,
\begin{align}
\mb x^t = \mb x^{t-1} - \eta^t \left( \mb H^{t-1}\right)^{-1} \pd{G^{t-1}\left(\mb x^{t-1} \right)}{\mb x}
,
\label{eq newton step}
\end{align}
where $\mb H^{t-1}$ is the Hessian of $G^{t-1}\left(\mb x\right)$.  The step length $\eta^t$ is typically unity, and will be discussed in Section \ref{sec bad up}.
\item \label{update_subfunc} Choose an index $j \in \{1...N\}$, and update the corresponding approximating subfunction $g_i^{t}\left( \mb x \right)$ using a second order power series around $\mb x^t$, while leaving all other subfunctions unchanged,
\nobreak \end{enumerate} \nobreak
\begin{align}
\label{eq subf upd}
g_i^t\left( \mb x \right)
 =
	\left\{\begin{array}{lcrl}
g_i^{t-1}\left( \mb x \right) & & i\neq j  \\ \\
\left[
	\begin{array}{l}
		f_i\left( \mb x^t \right)  \\ \quad
		+ \left( \mb x - \mb x^t \right)^T {f'_i\left( \mb x^t \right)} \\ \quad
		+ \frac{1}{2} \left( \mb x - \mb x^t \right)^T 
			\mb H_i^t \ 
			\left( \mb x - \mb x^t \right)
	\end{array}
\right]
  & & i = j & 
	\end{array}\right.
.
\end{align}
The constant and first order term in Equation \ref{eq subf upd} are set by evaluating the subfunction and gradient, $f_j\left(\mb x^t\right)$ and $f'_j\left(\mb x^t\right)$.  
The quadratic term $\mb H_j^t$ is set by using the BFGS %\cite{Dennis1977}
algorithm to generate an online approximation to the true Hessian %$\mb H_j^{true}$
of subfunction $j$ based on its history of gradient evaluations (see Section \ref{sec online hessian}).  
The Hessian of the summed approximating function $G^t\left(\mb x\right)$ in Equation \ref{eq newton step} is the sum of the Hessians for each $g^t_j\left(\mb x\right)$, $\mb H^t = \sum_j \mb H^t_j$.

\subsection{A Shared, Adaptive, Low-Dimensional Representation}
\label{sec subspace}
The dimensionality $M$ of $\mb x \in \mc R^M$ is typically large.  As a result, the memory and computational cost of working directly with the matrices $\mb H_i^t \in \mc R^{M\times M}$ is typically prohibitive, as is the cost of storing the history terms $\Delta {f'}$ and $\Delta \mb x$ required by BFGS (see Section \ref{sec online hessian}).  To reduce the dimensionality from $M$ to a tractable value, all history is instead stored and all updates computed in a lower dimensional subspace, with dimensionality between $K_{min}$ and $K_{max}$.  This subspace is constructed such that it includes the most recent gradient and position for every subfunction, and thus $K_{min} \geq 2N$.  This guarantees that the subspace includes both the steepest gradient descent direction over the full batch, and the update directions from the most recent Newton steps (Equation \ref{eq newton step}). %, which are likely to be the most productive direction of travel.

For the results in this paper, $K_{min} = 2 N$ and $K_{max} = 3 N$.  The subspace is represented by the orthonormal columns of a matrix $\mb P^t \in \mc R^{M \times K^t}$, $\left(\mb P^t\right)^T \mb P^t = \mb I$.  $K^t$ is the subspace dimensionality at optimization step $t$.

%TODO change delta notation?

\subsubsection{Expanding the Subspace with a New Observation}

At each optimization step, an additional column is added to the subspace, expanding it to include the most recent gradient direction.  This is done by first finding the component in the gradient vector which lies outside the existing subspace, and then appending that component to the current subspace,
\begin{align}
\mb q_\text{orth} &= f'_j\left( \mb x^t \right) - \mb P^{t-1} \left(\mb P^{t-1}\right)^T f'_j\left( \mb x^t \right), \\
\label{eq subfunction append}
\mb P^{t} &= \left[ \mb P^{t-1}\quad \frac{\mb q_\text{orth}}{\left|\left| \mb q_\text{orth} \right| \right|} \right]
,
\end{align}
where $j$ is the subfunction updated at time $t$.  The new position $\mb x^t$ is included automatically, since the position update was computed within the subspace $\mb P^{t-1}$.  Vectors embedded in the subspace $\mb P^{t-1}$ can be updated to lie in $\mb P^t$ simply by appending a $0$, since the first $K^{t-1}$ dimensions of $\mb P^t$ consist of $\mb P^{t-1}$.

\subsubsection{Restricting the Size of the Subspace}
\label{sec subspace collapse}
In order to prevent the dimensionality $K^t$ of the subspace from growing too large, whenever $K^t > K_{max}$, the subspace is collapsed to only include the most recent gradient and position measurements from each subfunction.  The orthonormal matrix representing this collapsed subspace is computed by a QR decomposition on the most recent gradients and positions.  A new collapsed subspace is thus computed as,
\begin{align}
\mb P' &= \text{orth}\left(\left[
	f'_1\left( \mb x^{\tau_1^t} \right) \cdots f'_N\left( \mb x^{\tau_N^t} \right) \quad
	 \mb x^{\tau_1^t} \cdots  \mb x^{\tau_N^t}
\right]\right)
\label{eq subspace collapse}
,
\end{align}
where $\tau_i^t$ indicates the learning step at which the $i$th subfunction was most recently evaluated, prior to the current learning step $t$.  
Vectors embedded in the prior subspace $\mb P$ are projected into the new subspace $\mb P'$ by multiplication with a projection matrix $\mb T = \left(\mb P'\right)^T \mb P$.  Vector components which point outside the subspace defined by the most recent positions and gradients are lost in this projection.

Note that the subspace $\mb P'$ lies within the subspace $\mb P$.  The QR decomposition and the projection matrix $\mb T$ are thus both computed within $\mb P$, reducing the computational and memory cost (see Section \ref{sec comp cost}).

\subsection{Online Hessian Approximation}
\label{sec online hessian}

An independent online Hessian approximation $\mb H_j^t$ is maintained for each subfunction $j$.  It is computed by performing BFGS on the history of gradient evaluations and positions for that single subfunction\footnote{We additionally experimented with Symmetric Rank 1 \cite{Dennis1977} updates to the approximate Hessian, but found they performed worse than BFGS.  See Supplemental Figure \ref{fig design choices}.}.

\subsubsection{History Matrices}

For each subfunction $j$, we construct two matrices, $\Delta {f'}$ and $\Delta \mb x$.  Each column of $\Delta {f'}$ holds the change in the gradient of subfunction $j$ between successive evaluations of that subfunction, including all evaluations up until the present time.  Each column of $\Delta \mb x$ holds the corresponding change in the position $\mb x$ between successive evaluations.  Both matrices are truncated after a number of columns $L$, meaning that they include information from only the prior $L+1$ gradient evaluations for each subfunction.  For all results in this paper, $L=10$ (identical to the default history length for the LBFGS implementation used in Section \ref{sec results}).

\subsubsection{BFGS Updates}
\label{sec bfgs updates}
The BFGS algorithm functions by iterating through the columns in $\Delta {f'}$ and $\Delta \mb x$, from oldest to most recent.  Let $s$ be a column index, and $\mb B_s$ be the approximate Hessian for subfunction $j$ after processing column $s$.  For each $s$, the approximate Hessian matrix $\mb B_s$ is set so that it obeys the secant equation $\Delta {f'}_s = \mb B_s \Delta \mb x_s$, where $\Delta {f'}_s$ and $\Delta \mb x_s$ are taken to refer to the $s$th columns of the gradient difference and position difference matrix respectively.

In addition to satisfying the secant equation, $\mb B_s$ is chosen such that the difference between it and the prior estimate $\mb B_{s-1}$ has the smallest weighted Frobenius norm\footnote{The weighted Frobenius norm is defined as $\left| \left| \mb E \right| \right|_{F, \mb W} = \left| \left| \mb W \mb E \mb W \right| \right|_{F}$.  For BFGS, $\mb W = \mb B_s^{-\frac{1}{2}}$ \cite{BFGShistory}.  Equivalently, in BFGS the unweighted Frobenius norm is minimized after performing a linear change of variables to map the new approximate Hessian to the identity matrix.}.  This produces the standard BFGS update equation
\begin{align}
\mb B_s
& =
\mb B_{s-1} + \frac{
			\Delta {f'}_s \Delta {f'}_s^T
			}{
			\Delta {f'}_s^T \Delta \mb x_s
			}
		- \frac{
			\mb B_{s-1} \Delta \mb x_s \Delta \mb x_s^T \mb B_{s-1} 
			}{
			\Delta \mb x_s^T \mb B_{s-1} \Delta \mb x_s
			}
.
\label{eq bfgs}
\end{align}
The final update is used as the approximate Hessian for subfunction $j$, $\mb H_j^t = \mb B_{\max(s)}$.

\begin{table}
\centering
%\begin{center}
\begin{tabular}{llll}
\em Optimizer & \em Computation per pass & \em Memory use \\
\hline \\ \vspace{-6mm} \\
SFO & $\mc O\left( QN + M N^2 \right)$  & $\mc O\left( MN \right)$  \\
SFO, `sweet spot' & $\mc O\left( QN \right)$  & $\mc O\left( MN \right)$  \\
LBFGS & $\mc O\left( QN  + ML \right)$  & $\mc O\left( M L \right)$  \\
SGD & $\mc O\left( QN \right)$  & $\mc O\left( M \right)$  \\
AdaGrad & $\mc O\left( QN \right)$  & $\mc O\left( M \right)$  \\
SAG & $\mc O\left( QN \right)$  & $\mc O\left( MN \right)$  \\
\end{tabular}
%\vspace{-2.5mm}
\caption{Leading terms in the computational cost and memory requirements for SFO and several competing algorithms.  $Q$ is the cost of computing the value and gradient for a single subfunction, $M$ is the number of data dimensions, $N$ is the number of subfunctions, and $L$ is the number of history terms retained.  ``SFO, `sweet spot''' refers to the case discussed in Section \ref{sec ideal} where the minibatch size is adjusted to match computational overhead to subfunction evaluation cost.  For this table, it is assumed that $M \gg N \gg L$.
%$L^{1.4} < N$, $L < MN$, $N < M$.
%*Minimize G currently implemented less efficiently than this
\label{tb cost compare}
}
%\end{center}
%\vspace{-3mm}
\end{table}
\begin{figure}[h]
\centering
%\begin{center}
\begin{tabular}{ccc}
\hspace{-5mm}
\begin{tabular}{c}
\hspace{-0.05\linewidth}
\includegraphics[width=0.52\linewidth]{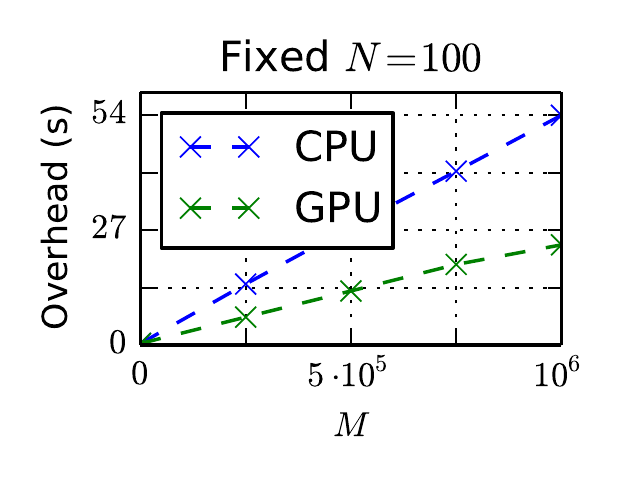} 
\hspace{-0.47\linewidth}(a)\hspace{0.47\linewidth}
\hspace{-0.06\linewidth}
\includegraphics[width=0.52\linewidth]{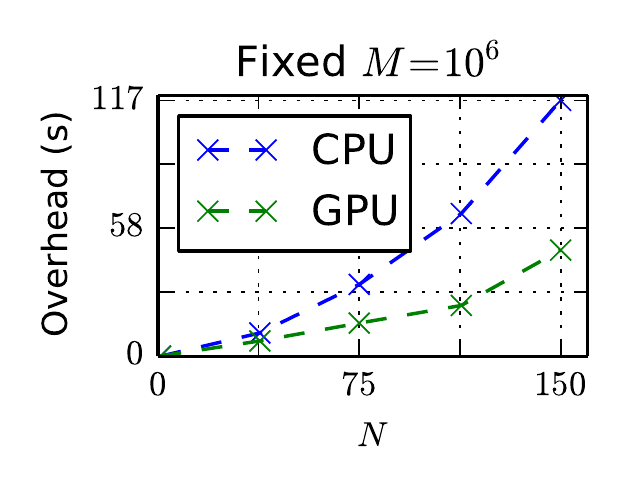}
\hspace{-0.47\linewidth}(b)\hspace{0.47\linewidth} 
\vspace{5mm}
\\
% 	\begin{tabular}{|c|}\hline
\hspace{-0.06\linewidth}
	 \includegraphics[width=0.45\linewidth]{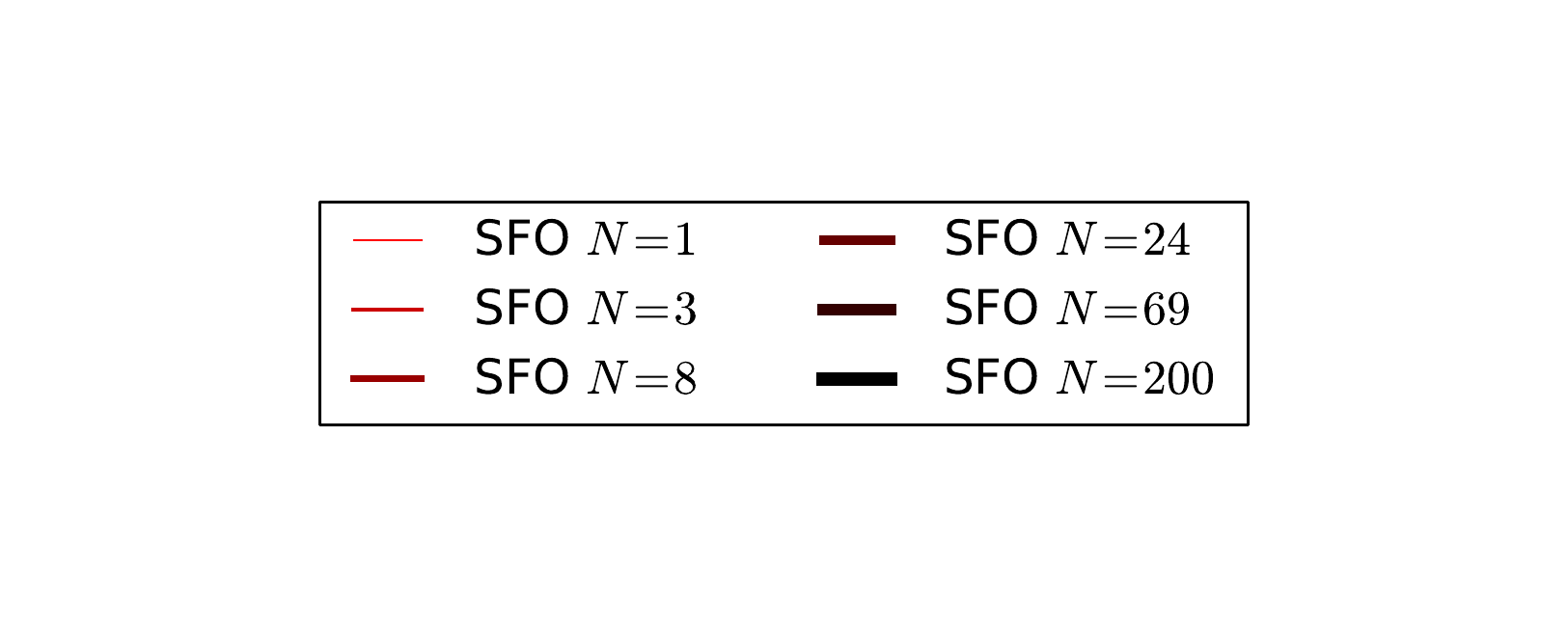} \\
\hspace{-0.06\linewidth}
	 \includegraphics[width=0.49\linewidth]{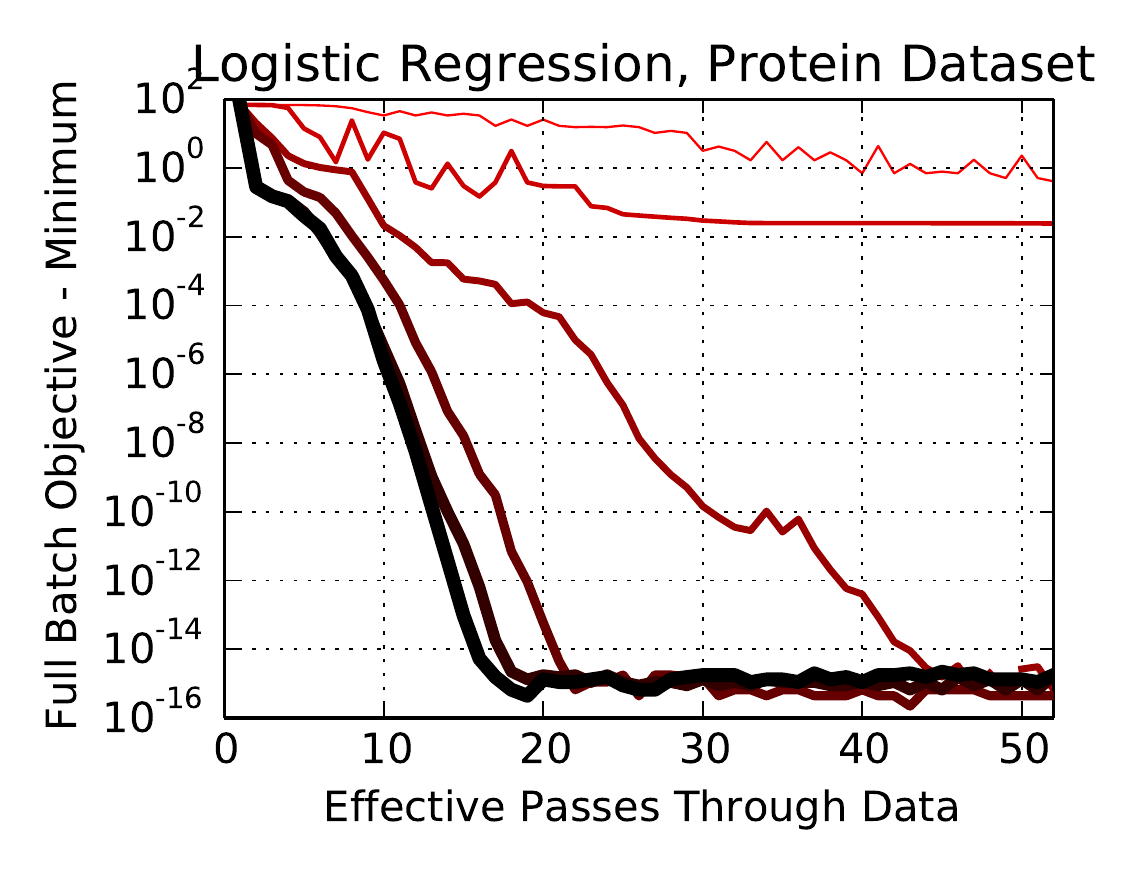}
	 \includegraphics[width=0.49\linewidth]{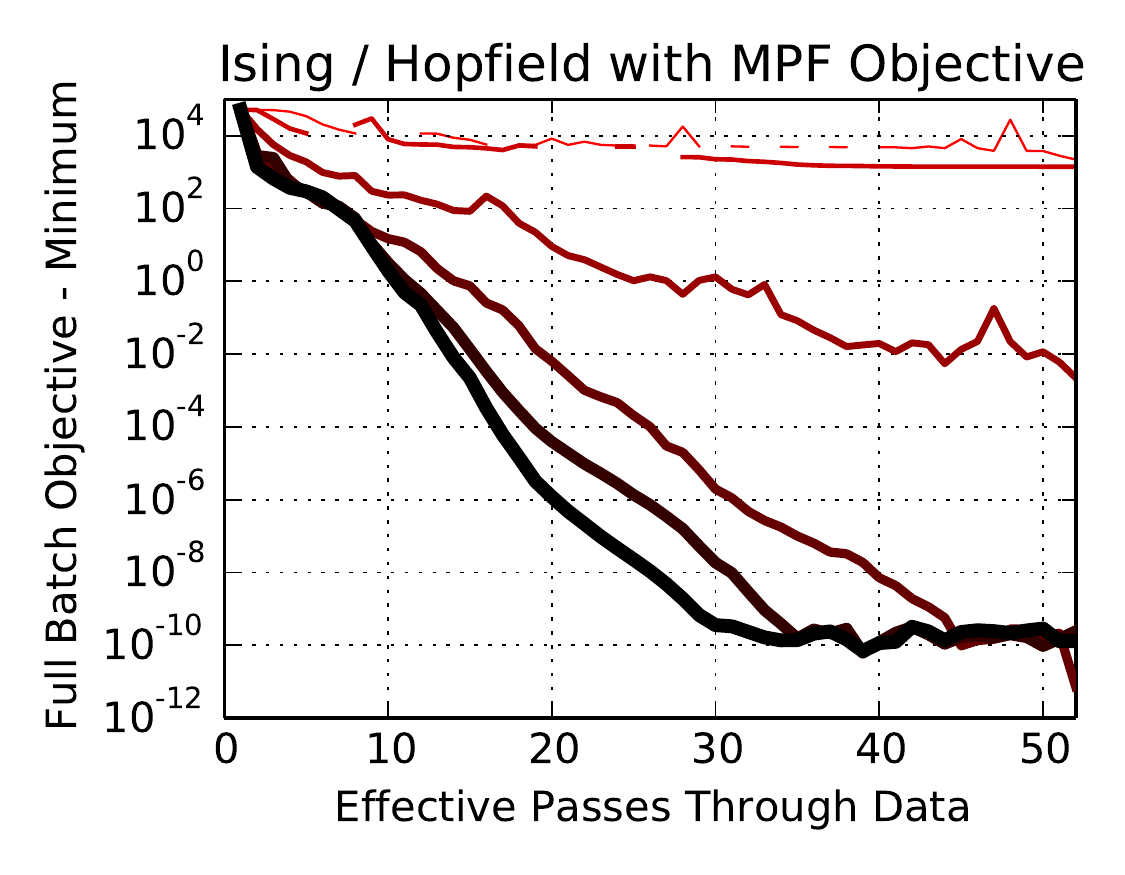}
%	\hline \end{tabular}	
\hspace{-0.99\linewidth}(c)\hspace{0.99\linewidth}	
\end{tabular}
\hspace{-5mm}
\end{tabular}
%\end{center}
%\vspace{-2.5mm}
\caption{
An exploration of computational overhead and optimizer performance, especially as the number of minibatches or subfunctions $N$ is adjusted. 
%Choosing the number of subfunctions.  
{\em (a)} Computational overhead required by SFO for a full pass through all the subfunctions as a function of dimensionality $M$ for fixed $N=100$.  {\em (b)} Computational overhead of SFO as a function of $N$ for fixed $M=10^6$.  Both plots show the computational time required for a full pass of the optimizer, excluding time spent computing the target objective and gradient.  This time is dominated by the $\mc O\left( M N^2 \right)$ cost per pass of $N$ iterations of subspace projection. % (see Supplemental Table \ref{tb cost}). 
{\em CPU} indicates that all computations were performed on a 2012 Intel i7-3970X CPU (6 cores, 3.5 GHz).  
{\em GPU} indicates that subspace projection was performed on a GeForce GTX 660 Ti GPU.  
{\em (c)} Optimization performance on the two convex problems in Section \ref{sec results} as a function of the number of minibatches $N$.  Note that near maximal performance is achieved after breaking the target problem into only a small number of minibatches.
}
\label{fig overhead}
%\vspace{-2.5mm}
\end{figure}
\section{Implementation Details}\label{sec imp det}

Here we briefly review additional design choices that were made when implementing this algorithm.  Each of these choices is presented more thoroughly in Appendix \ref{app implementation}.  Supplemental Figure \ref{fig design choices} demonstrates that the optimizer performance is robust to changes in several of these design choices.

\subsection{BFGS Initialization}
\label{sec bfgs init}

The first time a subfunction is evaluated (before there is sufficient history to run BFGS), the approximate Hessian $\mb H_j^t$ is set to the identity times the median eigenvalue of the average Hessian of the other active subfunctions.  For later evaluations, the initial BFGS matrix is set to a scaled identity matrix, $\mb B_0 = \beta \mb I$, where $\beta$ is the minimum eigenvalue found by solving the squared secant equation for the full history.  See Appendix \ref{app bfgs init} for details and motivation.

\subsection{Enforcing Positive Definiteness}

It is typical in quasi-Newton techniques to enforce that the Hessian approximation remain positive definite.  In SFO each $\mb H^t_i$ is constrained to be positive definite by an explicit eigendecomposition and setting any too-small eigenvalues to the median positive eigenvalue.  This is computationally cheap due to the shared low dimensional subspace (Section \ref{sec subspace}).  This is described in detail in Appendix \ref{app pos def}.

\subsection{Choosing a Target Subfunction}\label{sec order}

The subfunction $j$ to update in Equation \ref{eq subf upd} is chosen to be the one farthest from the current location $\mb x^t$, using the current Hessian approximation as the metric.  This is described more formally in Appendix \ref{app subf choose}.  As illustrated in Supplemental Figure \ref{fig design choices}, this distance based choice outperforms the commonly used random choice of subfunction.

\subsection{Growing the Number of Active Subfunctions}
\label{sec active growth}

For many problems of the form in Equation \ref{eq F}, the gradient information is nearly identical between the different subfunctions early in learning.  We therefore begin with only two active subfunctions, and expand the active set whenever the length of the standard error in the gradient across subfunctions exceeds the length of the gradient.  This process is described in detail in Appendix \ref{app grow active}.  As illustrated in Supplemental Figure \ref{fig design choices}, performance only differs from the case where all subfunctions are initially active for the first several optimization passes.

\subsection{Detecting Bad Updates}
\label{sec bad up}

Small eigenvalues in the Hessian can cause update steps to overshoot severely (ie, if higher than second order terms come to dominate within a distance which is shorter than the suggested update step).  It is therefore typical in quasi-Newton methods such as BFGS, LBFGS, and Hessian-free optimization to detect and reject bad proposed update steps, for instance by a line search.  In SFO, bad update steps are detected by comparing the measured subfunction value $f_j\left( \mb x^t \right)$ to its quadratic approximation $g_j^{t-1}\left( \mb x^t \right)$.  This is discussed in detail in Section \ref{app bad up}.

\section{Properties}

\subsection{Computational Overhead and Storage Cost}
\label{sec comp cost}
Table \ref{tb cost compare} compares the cost of SFO to competing algorithms.  The dominant computational costs are the $\mc O\left( MN \right)$ cost of projecting the $M$ dimensional gradient and current parameter values into and out of the $\mc O\left(N\right)$ dimensional active subspace for each learning iteration, and the $\mc O\left( Q \right)$ cost of evaluating a single subfunction.  The dominant memory cost is $\mc O\left( MN \right)$, and stems from storing the active subspace $\mb P^t$.  Table \ref{tb cost} in the Supplemental Material provides the contribution to the computational cost of each component of SFO.  Figure \ref{fig overhead} plots the computational overhead per a full pass through all the subfunctions associated with SFO as a function of $M$ and $N$.  If each of the $N$ subfunctions corresponds to a minibatch, then the computational overhead can be shrunk as described in Section \ref{sec ideal}.

Without the low dimensional subspace, the leading term in the computational cost of SFO would be the far larger $\mc O\left( M^{2.4} \right)$ cost per iteration of inverting the approximate Hessian matrix in the full $M$ dimensional parameter space, and the leading memory cost would be the far larger $\mc O\left( M^2N \right)$ from storing an $M\times M$ dimensional Hessian for all $N$ subfunctions.

\subsubsection{Ideal Minibatch Size}
\label{sec ideal}
Many objective functions consist of a sum over a number of data points $D$, where $D$ is often larger than $M$.  For example, $D$ could be the number of training samples in a supervised learning problem, or data points in maximum likelihood estimation.  To control the computational overhead of SFO in such a regime, it is useful to choose each subfunction in Equation \ref{eq approximating functions} to itself be a sum over a minibatch of data points of size $S$, yielding $N=\frac{D}{S}$.  This leads to a computational cost of evaluating a single subfunction and gradient of $\mc O\left(Q\right) = \mc O\left(MS\right)$.  The computational cost of projecting this gradient from the full space to the shared $N$ dimensional adaptive subspace, on the other hand, is $\mc O\left(MN\right) = \mc O\left(M\frac{D}{S}\right)$.  Therefore, in order for the costs of function evaluation and projection to be the same order, the minibatch size $S$ should be proportional to $\sqrt{D}$, yielding
\begin{align}
N &\propto \sqrt{D}
.
\end{align}  
The constant of proportionality should be chosen small enough that the majority of time is spent evaluating the subfunction.  In most problems of interest, $\sqrt{D} \ll M$, justifying the relevance of the regime in which the number of subfunctions $N$ is much less than the number of parameters $M$. Finally, the computational and memory costs of SFO are the same for sparse and non-sparse objective functions, while Q is often much smaller for a sparse objective.  Thus the ideal $S$ ($N$) is likely to be larger (smaller) for sparse objective functions.

\begin{figure*}%[htp]
%\vspace{-0.4in}
\centering
%\begin{center}
\begin{tabular}{ccc}
\hspace{-5mm}
\begin{tabular}{c}
\includegraphics[width=0.3\linewidth]{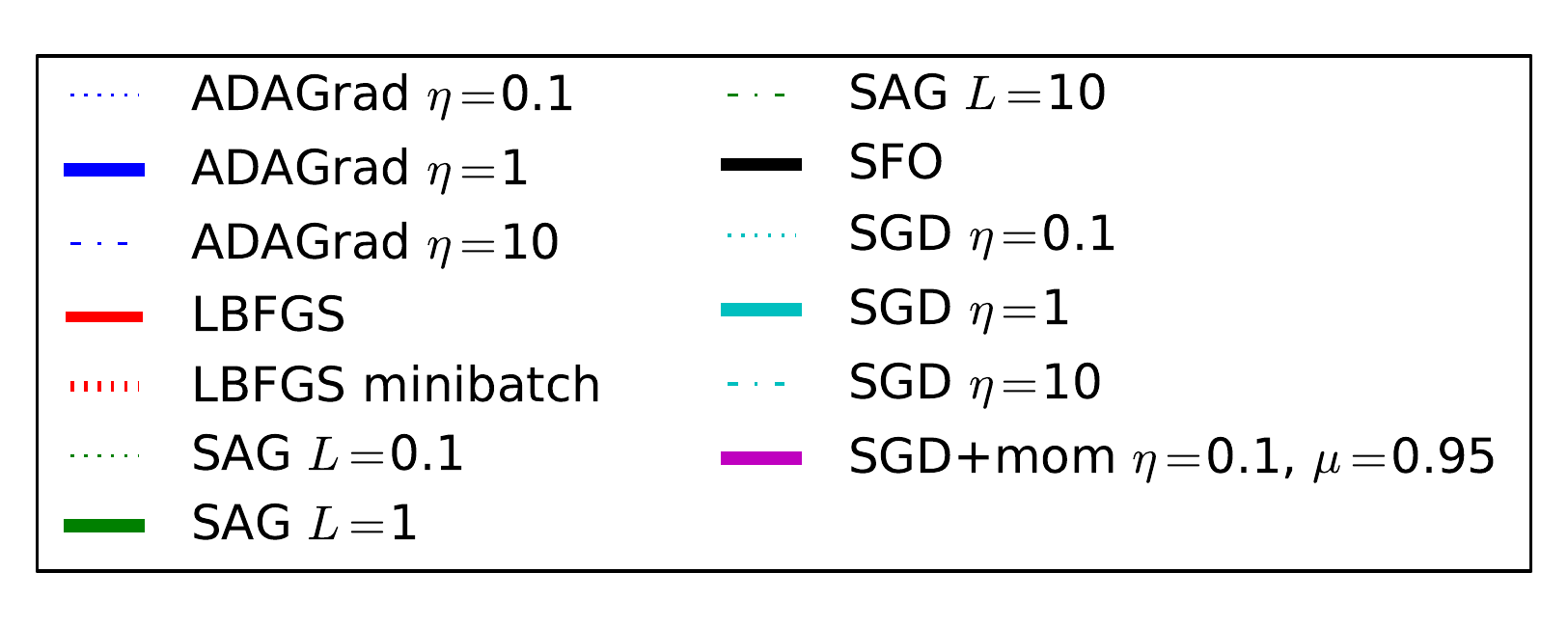} \\
(a)\includegraphics[width=0.3\linewidth]{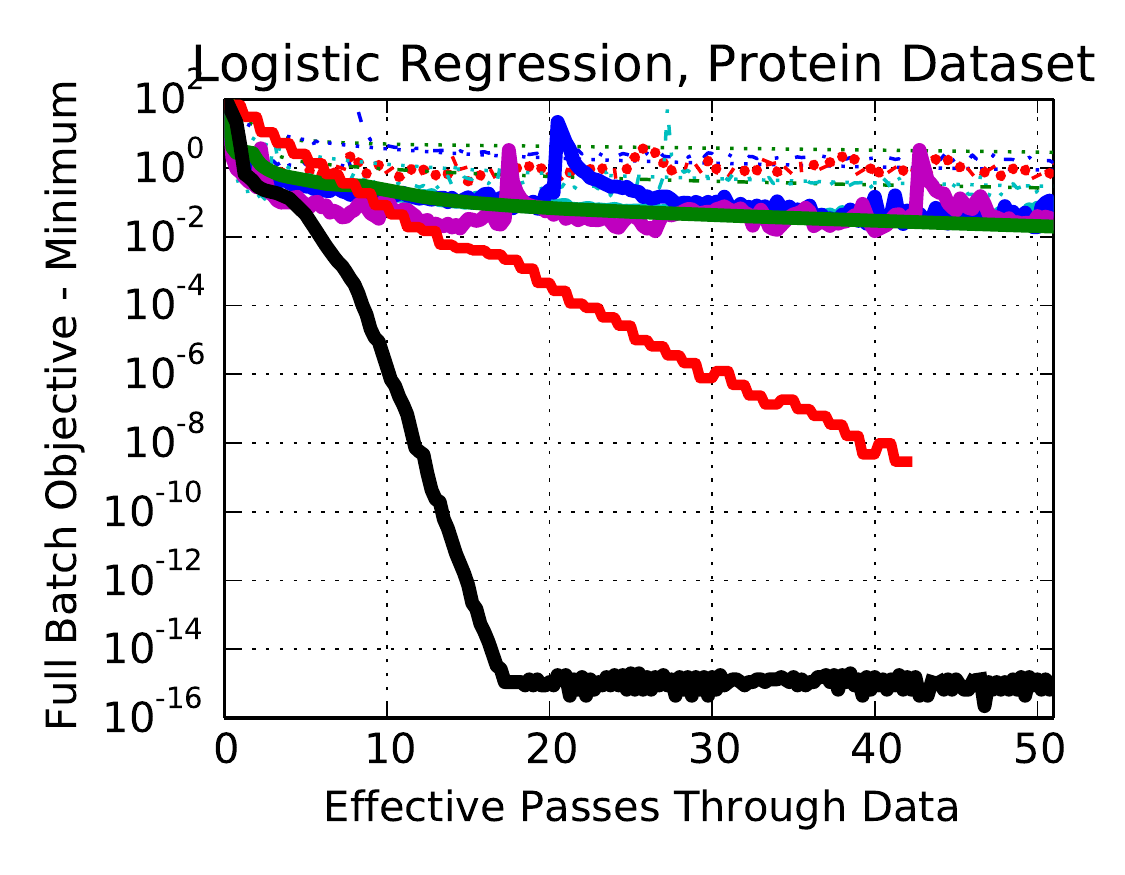}
\end{tabular}
\hspace{-5mm}
 & 
\begin{tabular}{c}
\includegraphics[width=0.3\linewidth]{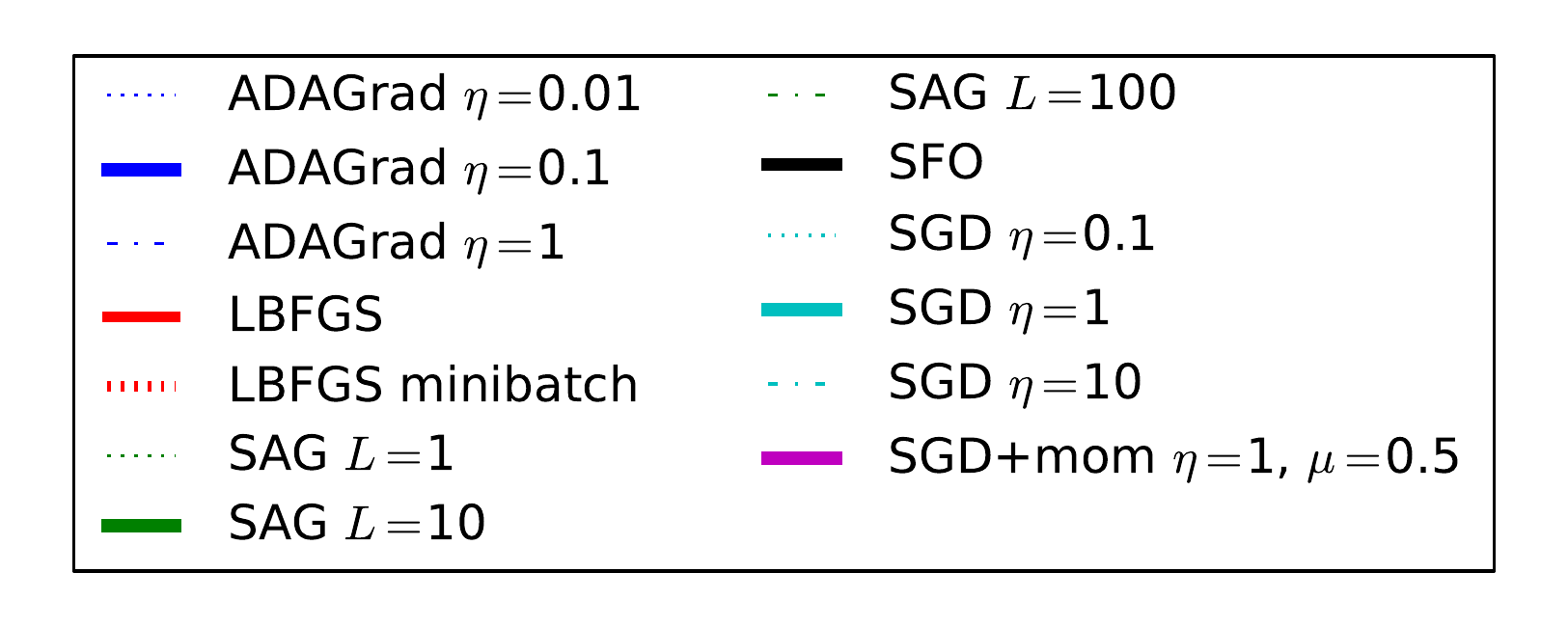} \\
(b)\includegraphics[width=0.3\linewidth]{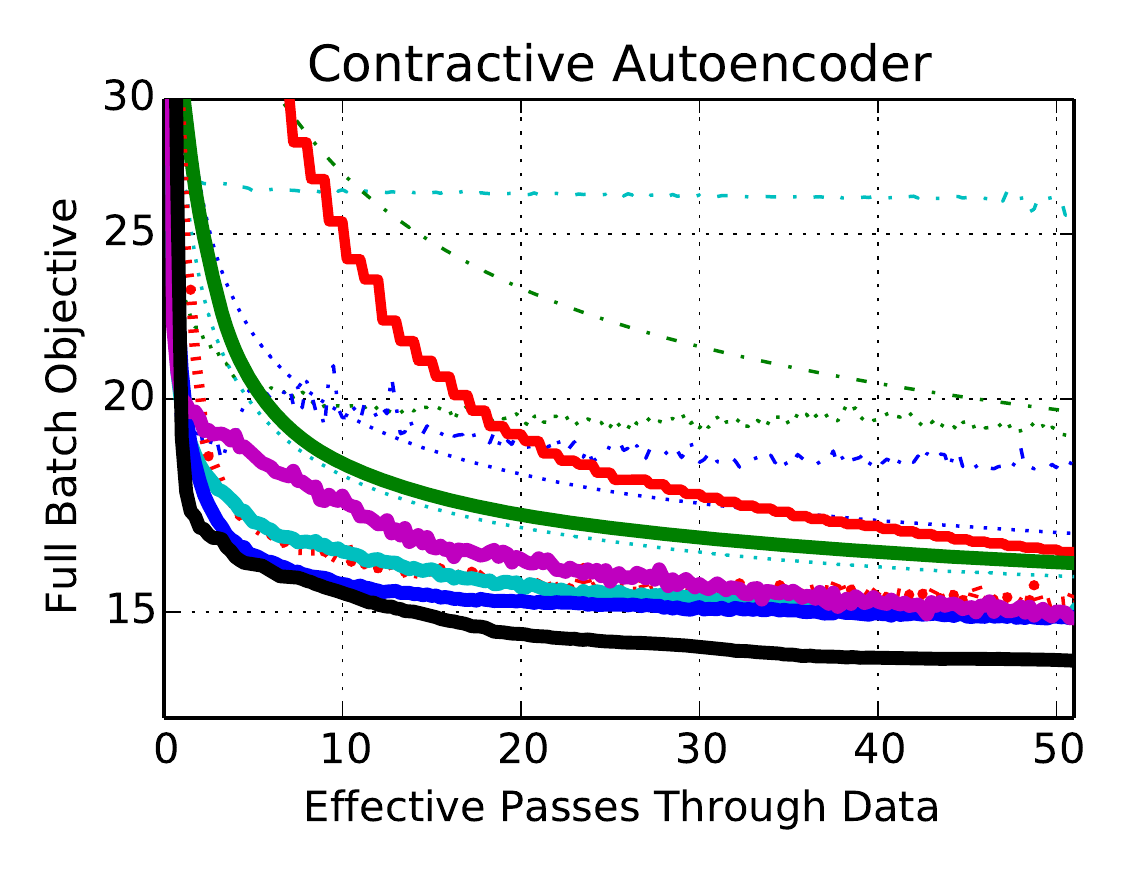}
\end{tabular}
\hspace{-5mm}
 & 
\begin{tabular}{c}
\includegraphics[width=0.3\linewidth]{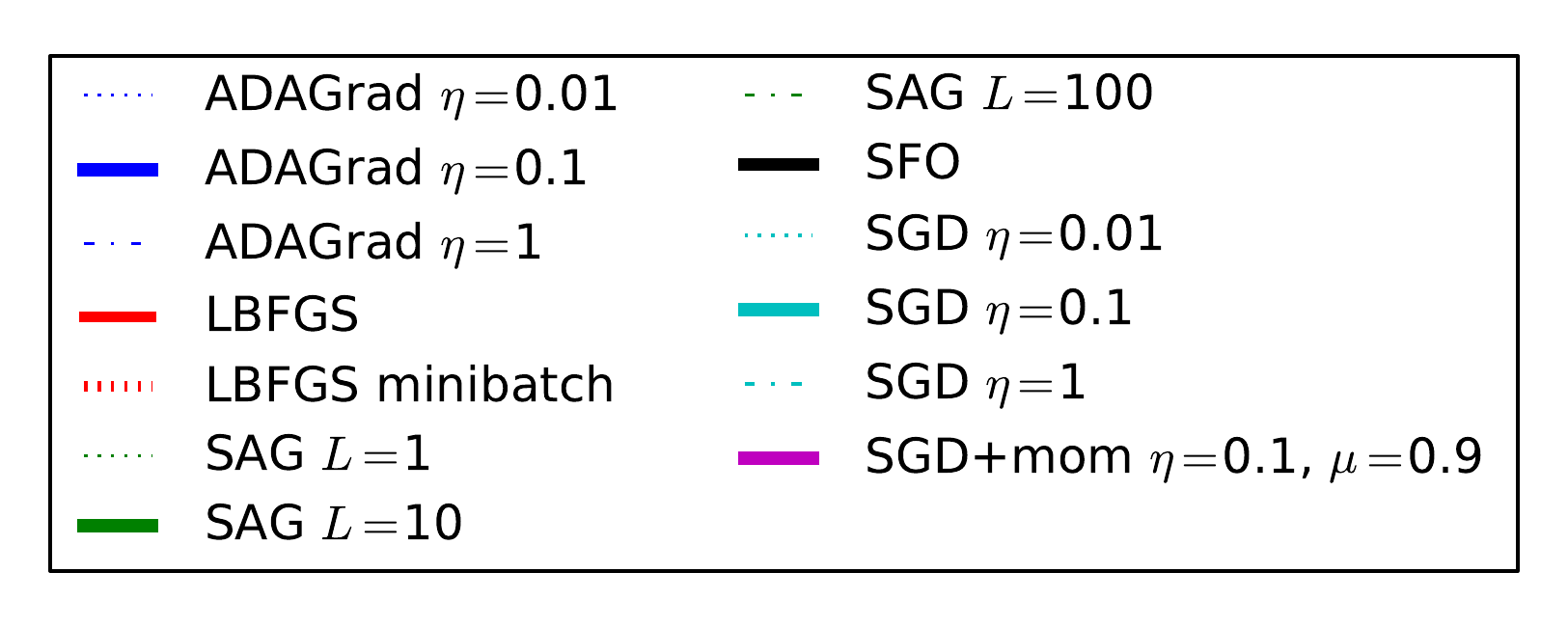} \\
(c)\includegraphics[width=0.3\linewidth]{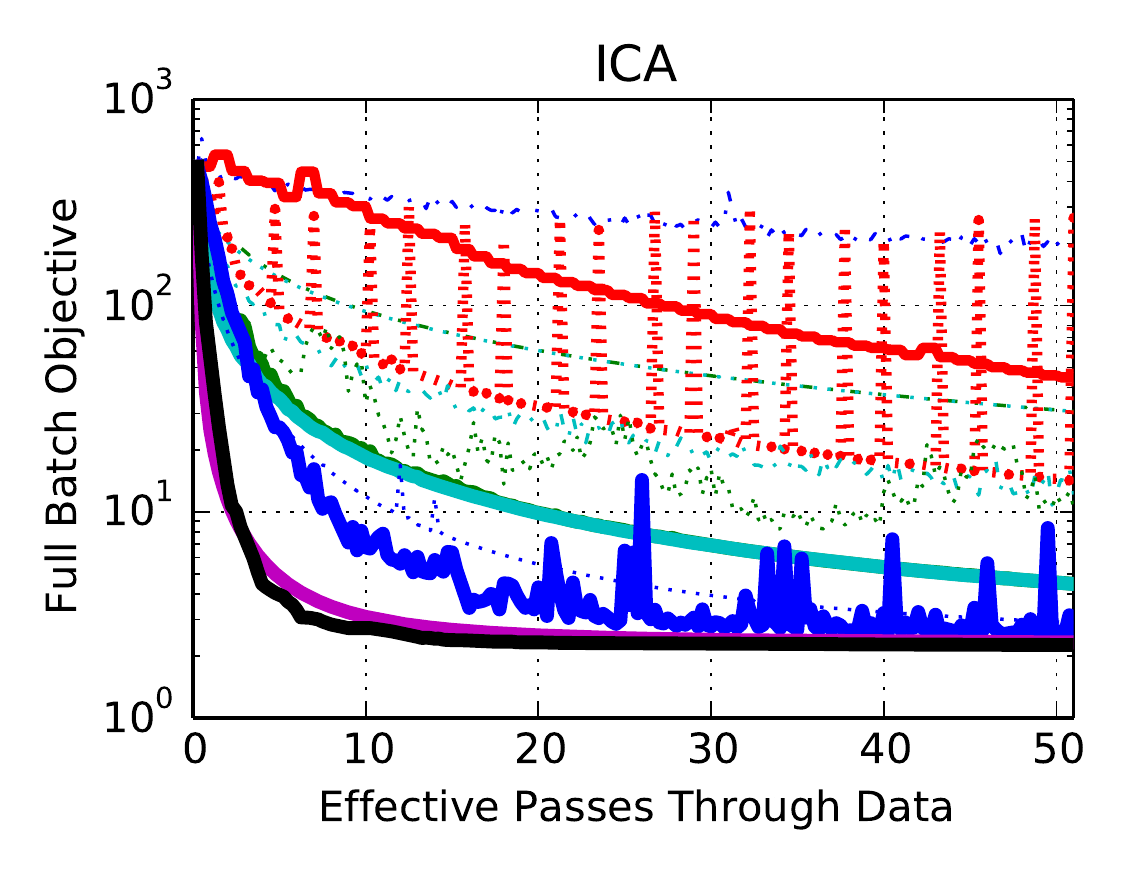}
\end{tabular}
\\ \\
\hspace{-5mm}
\begin{tabular}{c}
\includegraphics[width=0.30692307692307697\linewidth]{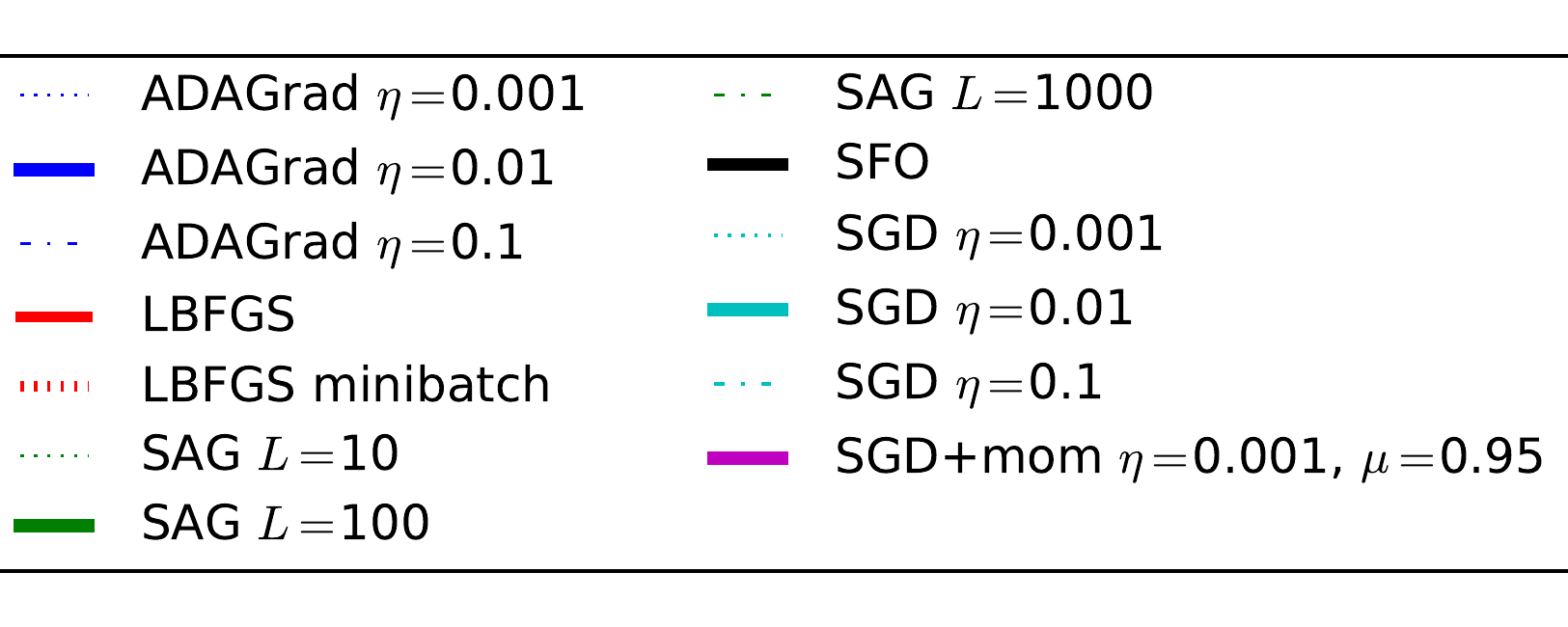} \\
(d)\includegraphics[width=0.3\linewidth]{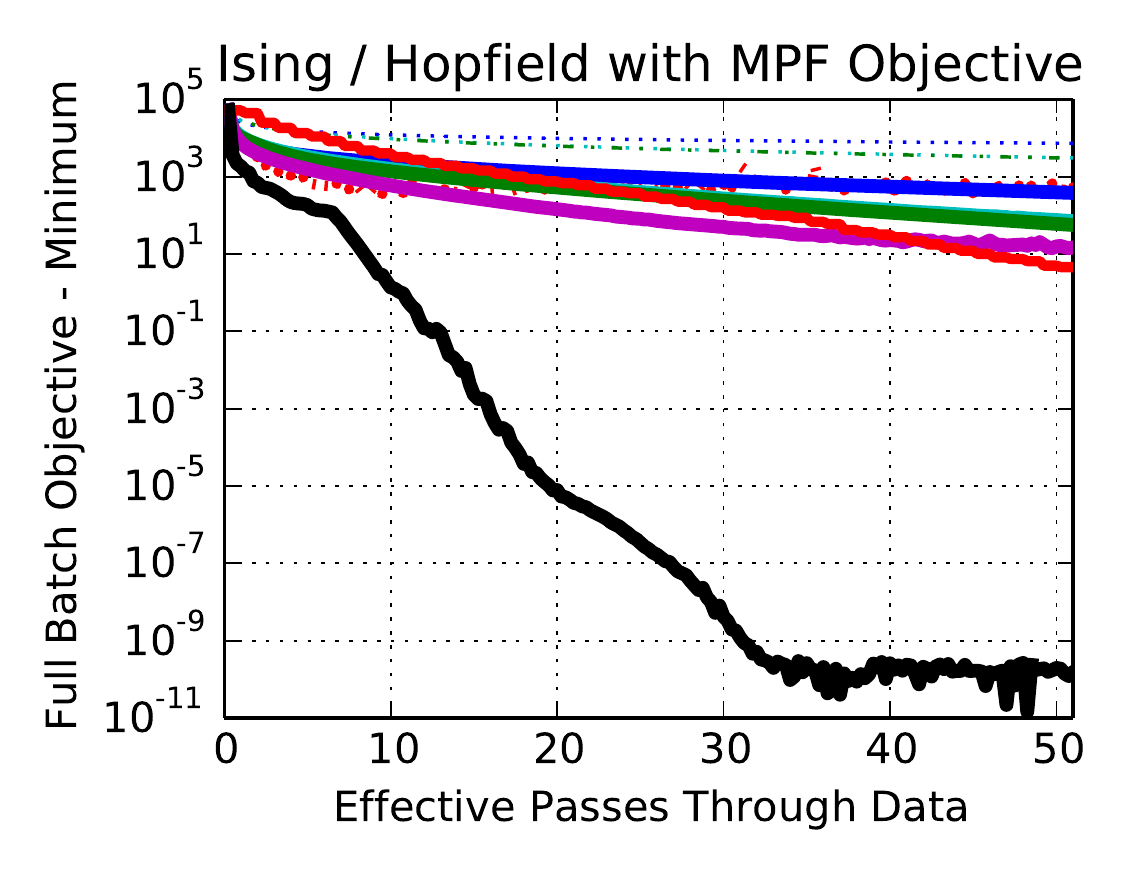}
\end{tabular}
\hspace{-5mm}
%\begin{tabular}{c}
%\includegraphics[width=0.3\linewidth]{figures/figure_MLP-hard_legend.pdf} \\
%\includegraphics[width=0.3\linewidth]{figures/figure_MLP-hard_true.pdf} \\
%(c)
%\end{tabular}
 & 
%\\ \\
\begin{tabular}{c}
\includegraphics[width=0.3\linewidth]{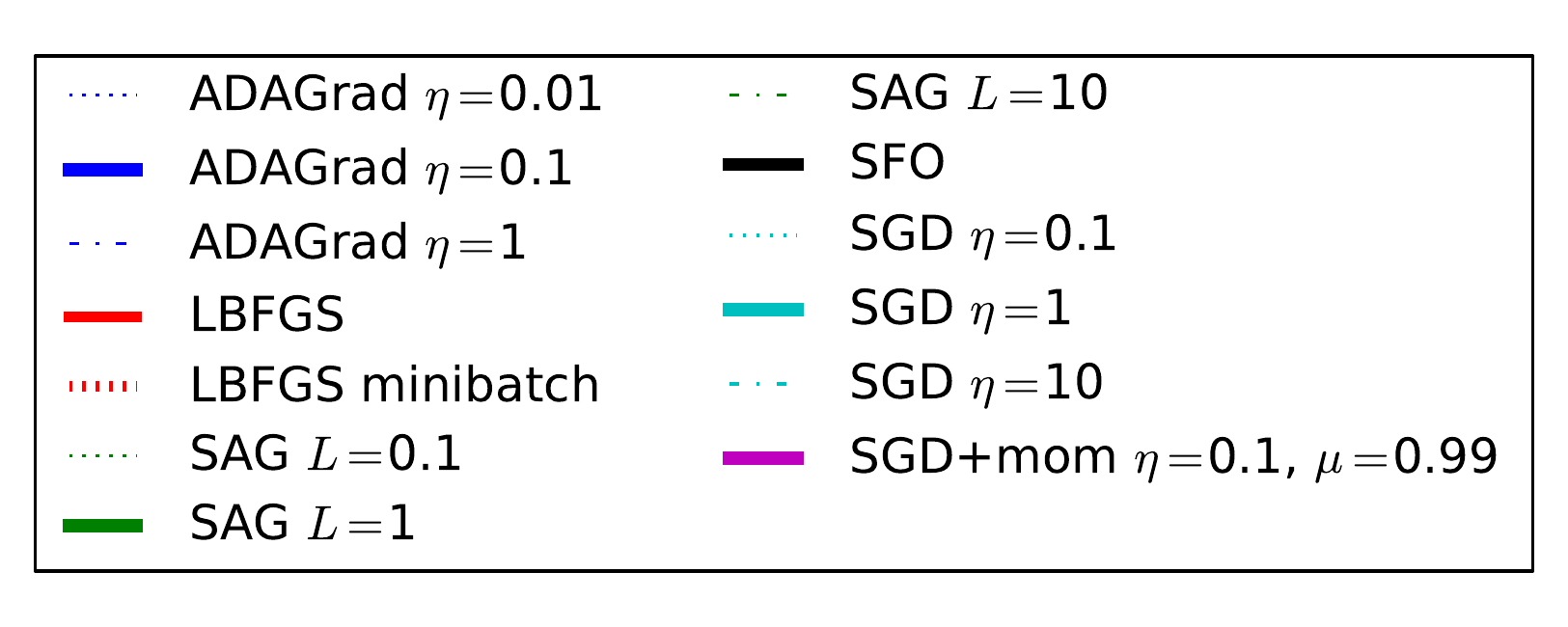} \\
(e)\includegraphics[width=0.3\linewidth]{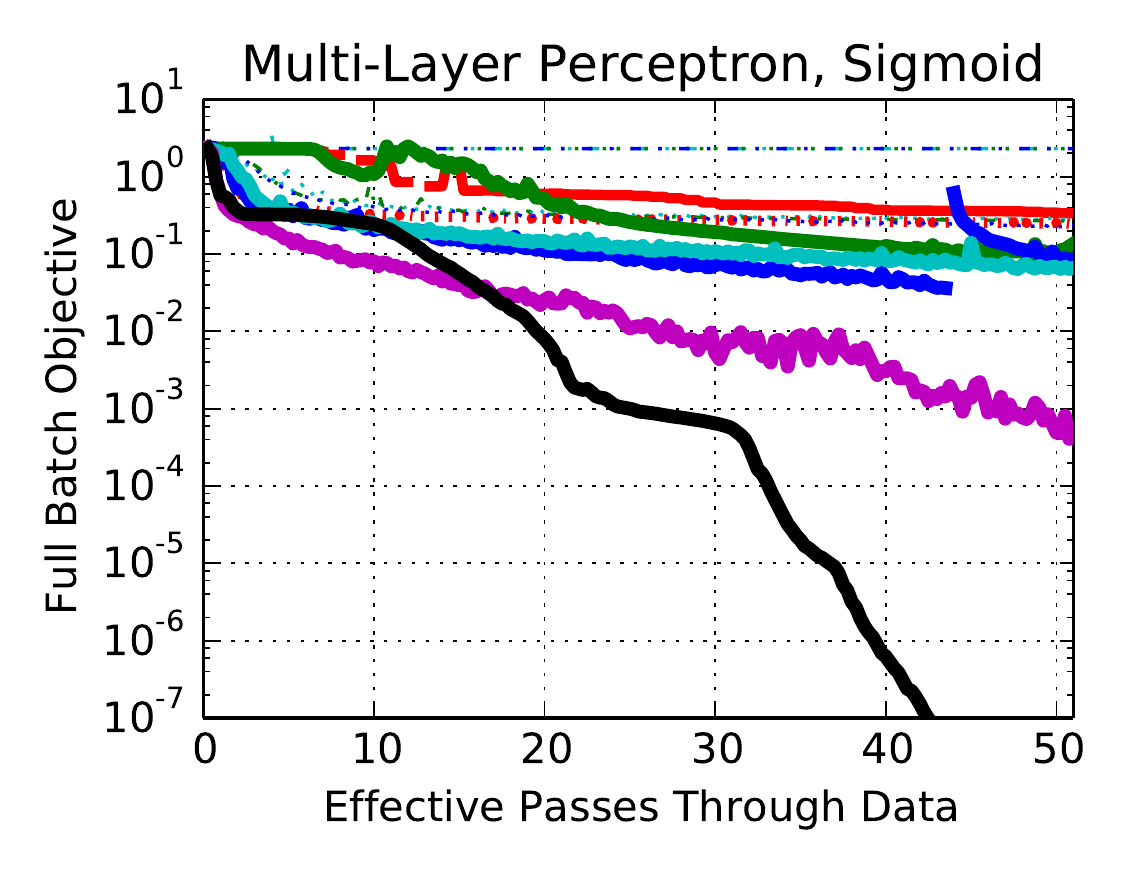}
\end{tabular}
\hspace{-5mm}
 & 
\begin{tabular}{c}
\includegraphics[width=0.3\linewidth]{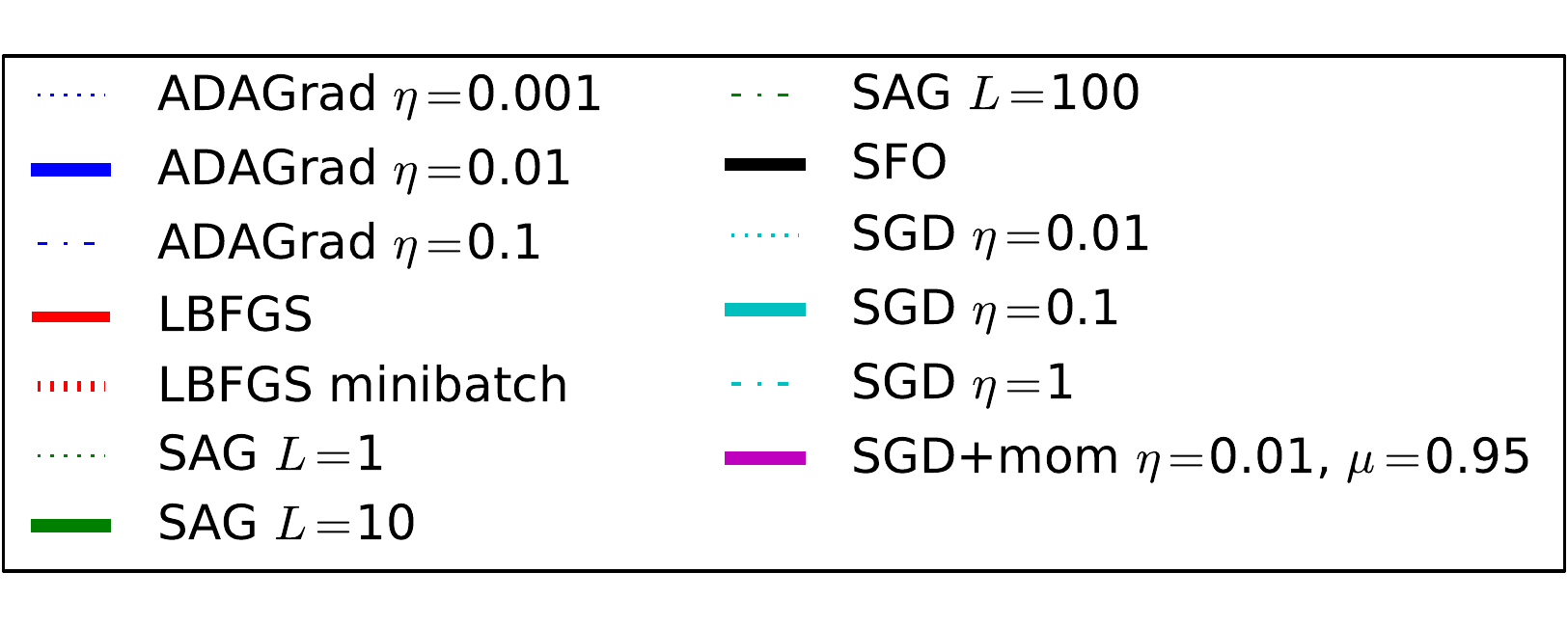} \\
(f)\includegraphics[width=0.3\linewidth]{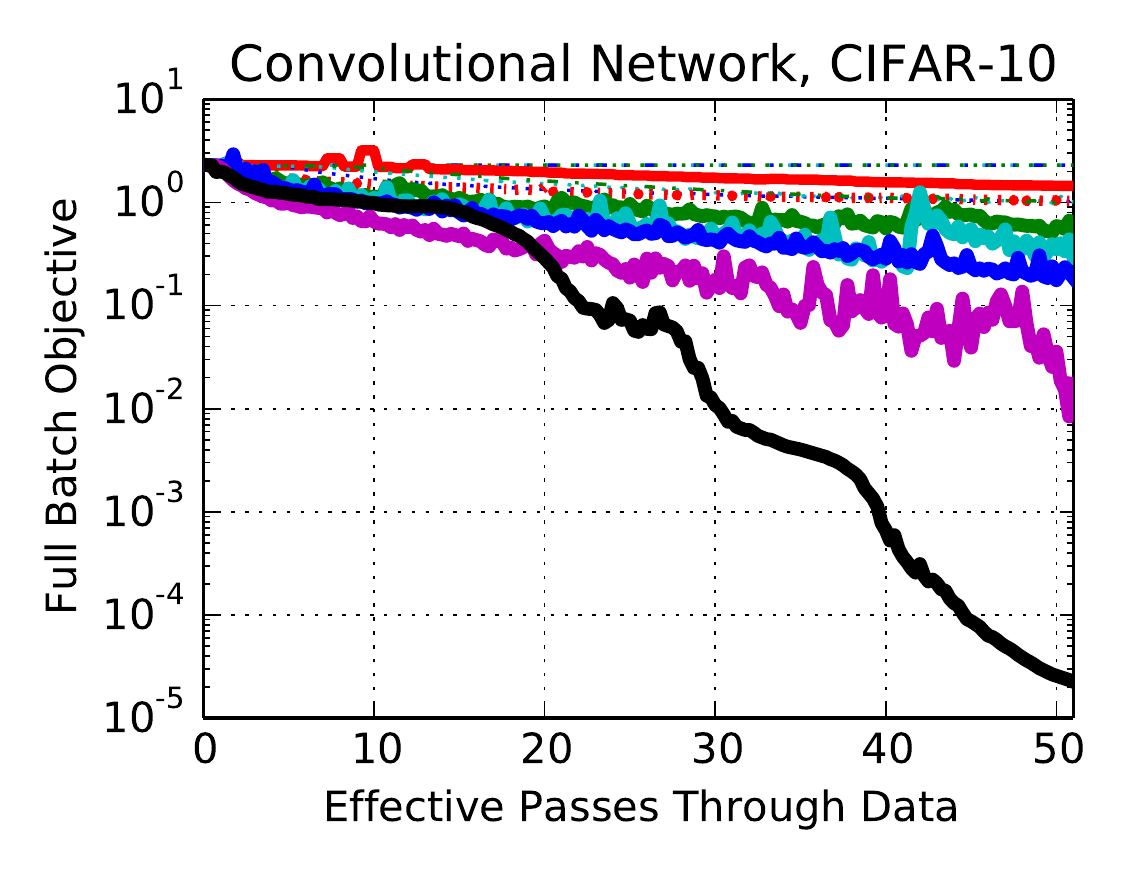}
\end{tabular}
\end{tabular}
%\end{center}
%\vspace{-2.5mm}
\caption{
A comparison of SFO to competing optimization techniques for six objective functions.  %The competing optimization techniques are described in Section \ref{sec results}.
The bold lines indicate the best performing hyperparameter for each optimizer.  Note that unlike all other techniques besides LBFGS, SFO does not require tuning hyperparameters (for instance, the displayed SGD+momentum traces are the best out of 32 hyperparameter configurations).  The objective functions shown are {\em (a)} a logistic regression problem, {\em (b)} a contractive autoencoder trained on MNIST digits, {\em (c)} an Independent Component Analysis (ICA) model trained on MNIST digits,
{\em (d)} an Ising model / Hopfield associative memory trained using Minimum Probability Flow,
%{\em (c)} a multi-layer perceptron with rectified linear units trained on MNIST digits, 
{\em (e)} a multi-layer perceptron with sigmoidal units trained on MNIST digits, and {\em (f)} a multilayer convolutional network with rectified linear units trained on CIFAR-10.  The logistic regression and MPF Ising objectives are convex, and their objective values are plotted relative to the global minimum.
}
\label{fig results}
%\vspace{-2.5mm}
\end{figure*}

Note that as illustrated in Figure \ref{fig overhead}c and Figure \ref{fig results} performance is very good even for small $N$.

\subsection{Convergence}

Concurrent work by \cite{Mairal2013} considers a similar algorithm to that described in Section \ref{sec update}, but with $\mb H_i^t$ a scalar constant rather than a matrix.  Proposition 6.1 in \cite{Mairal2013} shows that in the case that each $g_i$ majorizes its respective $f_i$, and subject to some additional smoothness constraints, $G^t\left(\mb x\right)$ monotonically decreases, and $\mb x^*$ is an asymptotic stationary point.  Proposition 6.2 in \cite{Mairal2013} further shows that for strongly convex $f_i$, the algorithm exhibits a linear convergence rate to $\mb x^*$.  A near identical proof should hold for a simplified version of SFO, with random subfunction update order, and with $\mb H_i^t$ regularized in order to guarantee satisfaction of the majorization condition.

\section{Experimental Results}
\label{sec exp results}
\label{sec results}
We compared our optimization technique to several competing optimization techniques for seven objective functions.  The results are illustrated in Figures \ref{fig results} and \ref{fig hf}, and the optimization techniques and objectives are described below.  For all problems our method outperformed all other techniques in the comparison.

Open source code which implements the proposed technique and all competing optimizers, and which directly generates the plots in Figures \ref{fig cartoon}, \ref{fig overhead}, and \ref{fig results}, is provided at \url{https://github.com/Sohl-Dickstein/Sum-of-Functions-Optimizer}.%\footnote{All figures in the paper can be reproduced simply by downloading code and training data, typing ``python figures.py'', and then waiting a week for all optimizers to run on all objective functions for all hyperparameters.}.

\subsection{Optimizers}
{\em SFO} refers to Sum of Functions Optimizer, and is the new algorithm presented in this paper.  
{\em SAG} refers to Stochastic Average Gradient method, with the trailing number providing the Lipschitz constant.  {\em SGD} refers to Stochastic Gradient Descent, with the trailing number indicating the step size.  {\em ADAGrad} indicates the AdaGrad algorithm, with the trailing number indicating the initial step size.  {\em LBFGS} refers to the limited memory BFGS algorithm.  {\em LBFGS minibatch} repeatedly chooses one tenth of the subfunctions, and runs LBFGS for ten iterations on them.  {\em Hessian-free} refers to Hessian-free optimization.

For {\em SAG}, {\em SGD}, and {\em ADAGrad} the hyperparameter was chosen by a grid search.  The best hyperparameter value, and the hyperparameter values immediately larger and smaller in the grid search, are shown in the plots and legends for each model in Figure \ref{fig results}.  In {\em SGD+momentum} the two hyperparameters for both step size and momentum coefficient were chosen by a grid search, but only the best parameter values are shown.  The grid-searched momenta were 0.5, 0.9, 0.95, and 0.99, and the grid-searched step lengths were all integer powers of ten between $10^{-5}$ and $10^2$.  For {\em Hessian-free}, the hyperparameters, source code, and objective function are identical to those used in \cite{Martens2010}, and the training data was divided into four ``chunks.''  For all other experiments and optimizers the training data was divided into $N = 100$ minibatches (or subfunctions).

%\subsection{Toy Problem}
%
%As a simple test problem, we constructed a sum of Euclidean norms, with random centers, raised to random powers.  The objective function is
%\begin{align}
%F\left( \mb x \right) &= \sum_i \left| \left| \mb x - \mb \mu_i \right| \right|^{\beta_i}
%,
%\end{align}
%where $\mu_i \sim N\left( \mb 0, \mb I \right)$, $\beta_i \sim U\left( 1, 4 \right)$, and $\mb x \in \mc R^{10}$.  This problem violates the second order Lipschitz continuity condition, which may be why the stochastic average gradient algorithm diverges for it.  A minimum value of 1 was assigned to $\beta_i$, because for smaller values the problem is not convex, and the gradient of $F\left( \mb x \right)$ goes to infinity near $\mu_i$.

\subsection{Objective Functions}

A detailed description of all target objective functions in Figure \ref{fig results} is included in Section \ref{sec supp objective} of the Supplemental Material.  In brief, they consisted of:
%\vspace{-2.5mm}
\begin{itemize}
  \item A logistic regression objective, chosen to be the same as one used in \cite{Roux2012}.
  \item A contractive autoencoder with 784 visible units, and 256 hidden units, similar to the one in \cite{Rifai2011}.
  \item An Independent Components Analysis (ICA) \cite{Bell1995} model with Student's t-distribution prior.
  \item An Ising model / Hopfield network trained using code from \cite{Hillar2012} implementing MPF \cite{MPF_ICML,SohlDickstein2011a}.
  \item A multilayer perceptron with a similar architecture to \cite{Hinton:2012tv}, with layer sizes of 784, 1200, 1200, and 10.  Training used Theano \cite{Bergstra2010}.
  \item A deep convolutional network with max pooling and rectified linear units, similar to \cite{Goodfellow2013}, with two convolutional layers with 48 and 128 units, and one fully connected layer with 240 units.  Training used Theano and Pylearn2 \cite{Goodfellow2013a}.
\end{itemize}

The logistic regression and Ising model / Hopfield objectives are convex, and are plotted relative to their global minimum.  The global minimum was taken to be the smallest value achieved on the objective by any optimizer.

\begin{figure}
\centering
\begin{tabular}{c}
\includegraphics[width=0.75\linewidth]{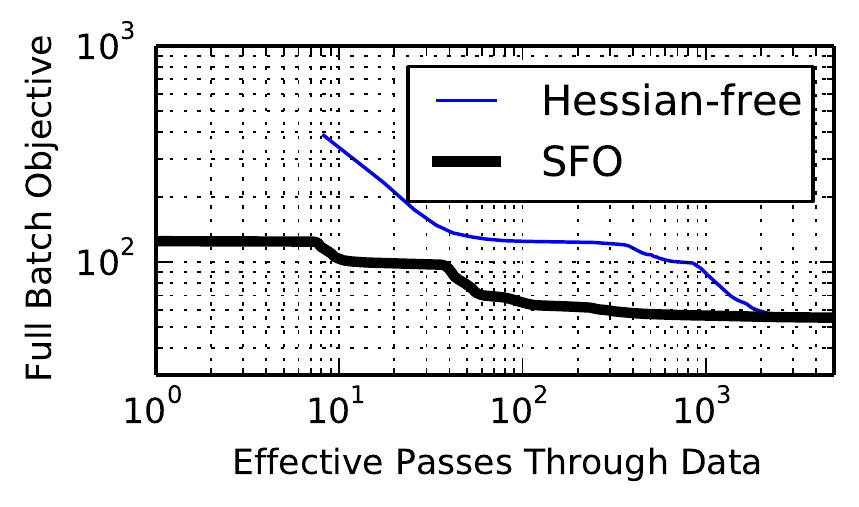}
\end{tabular}
%\vspace{-2.5mm}
\caption{
A comparison of SFO to Hessian-free optimization for a twelve layer neural network trained on the CURVES dataset. 
This problem is identical to an experiment in \cite{Martens2010}, and the Hessian-free convergence trace was generated using source code from the same paper. SFO converges in approximately one tenth the number of effective passes through the data as Hessian-free optimization.
}
%\vspace{-2.5mm}
\label{fig hf}
\end{figure}

In Figure \ref{fig hf}, a twelve layer neural network was trained on cross entropy reconstruction error for the CURVES dataset.  This objective, and the parameter initialization, was chosen to be identical to an experiment in \cite{Martens2010}.

\section{Future Directions}

We perform optimization in an $\mc O\left(N \right)$ dimensional subspace.  It may be possible, however, to drastically reduce the dimensionality of the active subspace without significantly reducing optimization performance.  For instance, the subspace could be determined by accumulating, in an online fashion, the leading eigenvectors of the covariance matrix of the gradients of the subfunctions, as well as the leading eigenvectors of the covariance matrix of the update steps.  This would reduce memory requirements and computational overhead even for large numbers of subfunctions (large $N$).

Most portions of the presented algorithm are naively parallelizable.  The $g_i^t\left( \mb x \right)$ functions can be updated asynchronously, and can even be updated using function and gradient evaluations from old positions $\mb x^\tau$, where $\tau < t$.  Developing a parallelized version of this algorithm could make it a useful tool for massive scale optimization problems.  Similarly, it may be possible to adapt this algorithm to an online / infinite data context by replacing subfunctions in a rolling fashion.

Quadratic functions are often a poor match to the geometry of the objective function \cite{Pascanu2012}.  Neither the dynamically updated subspace nor the use of independent approximating subfunctions $g_i^t\left(\mb x\right)$ which are fit to the true subfunctions $f_i\left(\mb x\right)$ depend on the functional form of $g_i^t\left(\mb x\right)$.  Exploring non-quadratic approximating subfunctions has the potential to greatly improve performance.

Section \ref{sec bfgs init} initializes the approximate Hessian using a diagonal matrix.  Instead, it might be effective to initialize the approximate Hessian for each subfunction using the average approximate Hessian from all other subfunctions.  Where individual subfunctions diverged they would overwrite this initialization.  This would take advantage of the fact that the Hessians for different subfunctions are very similar for many objective functions.

Recent work has explored the non-asymptotic convergence properties of stochastic optimization algorithms \cite{Moulines2011}.  It may be fruitful to pursue a similar analysis in the context of SFO.

Finally, the natural gradient \cite{Amari1998} can greatly accelerate optimization by removing the effect of dependencies and relative scalings between parameters.  The natural gradient can be simply combined with other optimization methods by performing a change of variables, such that in the new parameter space the natural gradient and the ordinary gradient are identical \cite{Sohl-Dickstein2012b}.  It should be straightforward to incorporate this change-of-variables technique into SFO.

\section{Conclusion}

We have presented an optimization technique which combines the benefits of LBFGS-style quasi-Newton optimization and stochastic gradient descent.  It does this by using BFGS to maintain an independent quadratic approximation for each contributing subfunction (or minibatch) in an objective function.  Each optimization step then alternates between descending the quadratic approximation of the full objective, and evaluating a single subfunction and updating the quadratic approximation for that single subfunction.  This procedure is made tractable in memory and computational time by working in a shared low dimensional subspace defined by the history of gradient evaluations.

%and then only evaluating a single subfunction .
%
% method for minibatch or sum of functions optimization which combines the
%
%TODO key advances table?  "independent hessian approximation for each subfunction" "adaptive low dimensional subspace is tractable in memory and time" 

\scriptsize
%\footnotesize
%\small
\bibliography{icml2014_sfo}
\bibliographystyle{icml2014}

\clearpage
\appendix

\normalsize

\icmltitle{Fast large-scale optimization by unifying stochastic gradient and quasi-Newton methods - Supplemental Material}

\section{Computational Complexity}
\label{sec cost appendix}
Here we provide a description of the computational cost of each component of the SFO algorithm.  See Table \ref{tb cost}.  The computational cost of matrix multiplication for $N\times N$ matrices is taken to be $\mc O\left(N^{2.4}\right)$.

\subsection{Function and Gradient Computation}

By definition, the cost of computing the function value and gradient for each subfunction is $\mc O\left( Q \right)$, and this must be done $N$ times to complete a full effective pass through all the subfunctions, yielding a total cost per pass of $\mc O\left( QN \right)$.

\subsection{Subspace Projection}

Once per iteration, the updated parameter values $\mb x^t$ must be projected from the $N$ dimensional adaptive low dimensional subspace into the full $M$ dimensional parameter space.  Similarly, once per iteration the gradient must be projected from the full $M$ dimensional parameter space into the $N$ dimensional subspace.  See Section \ref{sec subspace}.  Additionally the residual of the gradient projection must be appended to the subspace as described in Equation \ref{eq subfunction append}.  Each of these operations has cost $\mc O\left( MN \right)$, stemming from multiplication of a parameter or gradient vector by the subspace matrix $\mb P^t$.  They are performed $N$ times per effective pass through the data, yielding a total cost per pass of $\mc O\left( MN^2 \right)$.

\subsection{Subspace Collapse}

In order to constrain the dimensionality $K^t$ of the subspace to remain order $\mc O\left( N \right)$, the subspace must be collapsed every $\mc O\left( N \right)$ steps, or $\mc O\left( 1 \right)$ times per pass.  This is described in Section \ref{sec subspace collapse}.  The collapsed subspace is computed using a QR decomposition on the history terms (see Equation \ref{eq subspace collapse}) within the current subspace, with computational complexity $\mc O\left( N^3 \right)$.  The old subspace matrix $\mb P$ is then projected into the new subspace matrix $\mb P'$, involving the multiplication of a $\mc O\left( M \times N \right)$ matrix with a $\mc O\left( N \times N \right)$ projection matrix, with corresponding complexity $\mc O\left( MN^{1.4} \right)$.  The total complexity per pass is thus $O\left( MN^{1.4} + N^3 \right)$.

\subsection{Minimize $G^t\left(\mb x\right)$}

$G^t\left(\mb x\right)$ is minimized by an explicit matrix inverse in Equation \ref{eq newton step}.  Computing this inverse has cost $\mc O\left( N^{2.4} \right)$, and must be performed $N$ times per effective pass through the data.

With only small code changes, this inverse could instead be updated each iteration using the Woodbury identity and the inverse from the prior step, with cost $\mc O\left( N^{2} \right)$.  However, minimization of $G^t\left(\mb x\right)$ is not a leading contributor to the overall computational cost, and so increasing its efficiency would have little effect.

%Very good approximate solutions to systems of linear equations can typically be found with complexity $< \mc O\left( N^{2.4} \right)$, and a practical implementation of this component, not using a matrix inverse, could therefore be implemented with cost per iteration $< \mc O\left( N^{2.4} \right)$ and cost per pass $< \mc O\left( N^{3.4} \right)$.  However, minimization of $G^t\left(\mb x\right)$ is not a leading contributor to the overall computational cost, and so increasing its efficiency would have little effect.

\subsection{BFGS}

The BFGS iterations are performed in the $\mc O\left(L\right)$ dimensional subspace defined by the columns of $\Delta {f'}$ and $\Delta \mb x$.  Since BFGS consists of $L$ rank two updates of an $\mc O\left(L\times L\right)$ matrix, the cost of performing BFGS iterations is $\mc O\left( L^3 \right)$.  See Section \ref{sec bfgs updates}.  The cost of using a QR decomposition to compute the $L$ dimensional subspace defined by the columns of $\Delta {f'}$ and $\Delta \mb x$ is $\mc O\left( N L^2 \right)$, and the cost of projecting the $L$ history terms of length $N$ into the subspace is $\mc O\left( N L^{1.4} \right)$.  BFGS is performed $N$ times per effective pass through the data.  The total cost of BFGS is therefore $\mc O\left( N^2 L^2 + NL^3 \right)$.

In the current implementation, the full BFGS chain for a subfunction is recomputed every iteration.  However, BFGS could be modified to only compute the rank two update to the prior Hessian approximation at each iteration.  This would sacrifice the history-dependent initialization described in Section \ref{sec bfgs init}.  The resulting complexity per iteration would instead be $\mc O\left( N^2 \right)$, and the computational complexity per pass would instead be $\mc O\left( N^3 \right)$.  BFGS is also not a leading contributor to the overall computational cost, and so increasing its efficiency would have little effect.

\section{Objective Functions}
\label{sec supp objective}
A more detailed description of the objective functions used for experiments in the main text follows.

\subsection{Logistic Regression}
\label{sec logistic}
We chose the logistic regression objective, L2 regularization penalty, and training dataset to be identical to the protein homology test case in the recent Stochastic Average Gradient paper \cite{Roux2012}, to allow for direct comparison of techniques.  The one difference is that our total objective function is divided by the number of samples per minibatch, but unlike in \cite{Roux2012} is not also divided by the number of minibatches.  This different scaling places the hyperparameters for all optimizers in the same range as for our other experiments.

\subsection{Autoencoder}

We trained a contractive autoencoder, which penalizes the Frobenius norm of the Jacobian of the encoder function, on MNIST digits.  Autoencoders of this form have been successfully used for learning deep representations in neural networks \cite{Rifai2011}.  Sigmoid nonlinearities were used for both encoder and decoder.  The regularization penalty was set to 1, and did not depend on the number of hidden units.  The reconstruction error was divided by the number of training examples per minibatch.  There were 784 visible units, and 256 hidden units.

\subsection{Independent Components Analysis}
\label{sec ICA}
We trained an Independent Components Analysis (ICA) \cite{Bell1995} model with Student's t-distribution prior on MNIST digits by minimizing the negative log likelihood of the ICA model under the digit images.  Both the receptive fields and the Student's t shape parameter were estimated.  Digit images were preprocessed by performing PCA whitening and discarding components with variance less than $10^{-4}$ times the maximum variance.  The objective function was divided by the number of training examples per minibatch.

\subsection{Ising Model / Hopfield Network via MPF}

We trained an Ising/Hopfield model on MNIST digits, using code from \cite{Hillar2012}.  Optimal Hopfield network storage capacity can be achieved by training the corresponding Ising model via MPF \cite{Hillar2012,MPF_ICML,SohlDickstein2011a}.   The MPF objective was divided by the number of training examples per minibatch.  An L2 regularization penalty with coefficient 0.01 was added to the objective for each minibatch.

\subsection{Multilayer Perceptron}

We trained a deep neural network to classify digits on the MNIST digit recognition benchmark. We used a similar architecture to \cite{Hinton:2012tv}.  Our network consisted of: 784 input units, one hidden layer of 1200 units, a second hidden layer of 1200 units, and 10 output units. We ran the experiment using both rectified linear and sigmoidal units. 
The objective used was the standard softmax regression on the output units. Theano \cite{Bergstra2010} was used to implement the model architecture and compute the gradient.

\subsection{Deep Convolutional Network}

We trained a deep convolutional network on CIFAR-10 using max pooling and rectified linear units. The architecture we used contains two convolutional layers with 48 and 128 units respectively, followed by one fully connected layer of 240 units.  This architecture was loosely based on \cite{Goodfellow2013}.  Pylearn2 \cite{Goodfellow2013a} and Theano were used to implement the model.

\section{Implementation Details}\label{app implementation}

Here we expand on the implementation details which are outlined in Section \ref{sec imp det} in the main text.  Figure \ref{fig design choices} provides empirical motivation for several of the design choices made in this paper, by showing the change in convergence traces when those design choices are changed.  Note that even when these design choices are changed, convergence is still more rapid than for the competing techniques in Figure \ref{fig results}.

\subsection{BFGS Initialization}

\paragraph{No History}

An approximate Hessian can only be computed as described in Section \ref{sec online hessian} after multiple gradient evaluations.  If a subfunction $j$ only has one gradient evaluation, then its approximate Hessian $\mb H_j^t$ is set to the identity times the median eigenvalue of the average Hessian of the other active subfunctions.  If $j$ is the very first subfunction to be evaluated, $\mb H_j^t$ is initialized as the identity matrix times a large positive constant ($10^6$).

\paragraph{The First BFGS Step}\label{app bfgs init}

The initial approximate Hessian matrix used in BFGS is set to a scaled identity matrix, so that $\mb B_0 = \beta \mb I$.  This initialization will be overwritten by Equation \ref{eq bfgs} for all explored directions.  It's primary function, then, is to set the estimated Hessian for unexplored directions.  Gradient descent routines tend to progress from directions with large  slopes and curvatures, and correspondingly large eigenvalues, to directions with shallow slopes and curvatures, and smaller eigenvalues.  The typical eigenvalue in an unexplored direction is thus expected to be smaller than in previously explored directions.  We therefore set $\beta$ using a measure of the smallest eigenvalue in an explored direction

The scaling factor $\beta$ is set to the smallest non-zero eigenvalue of a matrix $\mb Q$,
%\begin{align}
$\beta = 
	\min_{\lambda_Q > 0} \lambda_Q$,
%\label{eq lambda}
%\end{align}
where $\lambda_Q$ indicates the eigenvalues of $\mb Q$.  
$\mb Q$ is the symmetric matrix with the smallest Frobenius norm which is consistent with the squared secant equations for all columns in $\Delta {f'}$ and $\Delta \mb x$.  That is,
\begin{align}
\mb Q & = 
	\left[
		{{\left(\Delta \mb x\right)}^+}^T
		{\left(\Delta {f'}\right)}^T
		\Delta {f'}
		{\left(\Delta \mb x\right)}^+
	\right]^\frac{1}{2}
\label{eq Q}
,
\end{align}
where $^+$ indicates the pseudoinverse, and $\frac{1}{2}$ indicates the matrix square root.  All of the eigenvalues of $\mb Q$ are non-negative.  $\mb Q$ and $\lambda_Q$ are computed in the subspace defined by $\Delta {f'}$ and $\Delta \mb x$, reducing computational cost (see Section \ref{sec comp cost}).

\subsection{Enforcing Positive Definiteness}\label{app pos def}

It is typical in quasi-Newton techniques to enforce that the Hessian approximation remain positive definite.  In SFO, at the end of the BFGS procedure, each $\mb H^t_i$ is constrained to be positive definite by performing an eigendecomposition, and setting any eigenvalues which are too small to the median positive eigenvalue.  The median is used because it provides a measure of ``typical'' curvature.  When an eigenvalue is negative (or {\em extremely} close to 0), it provides a poor estimate of curvature over the interval required to reach a minimum in the direction of the corresponding eigenvector.  Replacing it with the median eigenvalue therefore provides a more reasonable estimate.  If $\lambda_{max}$ is the maximum eigenvalue of $\mb H^t_i$, then any eigenvalues smaller than $\gamma \lambda_{max}$ are set to be equal to $\median_{\lambda>0} \lambda$.  For all experiments shown here, $\gamma = 10^{-8}$.  As described in Section \ref{sec subspace}, a shared low dimensional representation makes this eigenvalue computation tractable.

\subsection{Choosing a Target Subfunction}\label{app subf choose}

The subfunction $j$ to update in Equation \ref{eq subf upd} is chosen as,
\begin{align}
j &= \argmax_i \left[ \mb x^t - \mb x^{\tau_i} \right]^T \mb Q^t \left[ \mb x^t - \mb x^{\tau_i} \right]
,
\label{eq subf choose}
\end{align}
where $\tau_i$ indicates the time at which subfunction $i$ was last evaluated, and $\mb Q$ is either set to either $\mb H_i^t$ or $\mb H^t$, with probability $\frac{1}{2}$ for each.

That is, the updated subfunction is the one which was last evaluated farthest from the current location, using the approximate Hessian as a metric. This is motivated by the observation that the approximating functions which were computed farthest from the current location tend to be the functions which are least accurate at the current location, and therefore the most useful to update. The approximate Hessian $\mb H_i^t$ for the single subfunction is typically a more accurate measure of distance for that subfunction -- but we also use the approximate Hessian $\mb H^t$ for the full objective in order to avoid a bad Hessian estimate for a single subfunction preventing that subfunction from ever being evaluated.

This contrasts with the cyclic choice of subfunction in \cite{Blatt2007}, and the random choice of subfunction in \cite{Roux2012}.  See Supplemental Figure \ref{fig design choices} for a comparison of the optimization performance corresponding to each update ordering scheme.

\subsection{Growing the Number of Active Subfunctions}
\label{app grow active}

For many problems of the form in Equation \ref{eq F}, the gradient information is nearly identical between the different subfunctions early in learning.  We therefore begin with only a small number of active subfunctions, and expand the active set as learning progresses.  
We expand the active set by one subfunction every time the average gradient shrinks to within a factor $\alpha$ of the standard error in the average gradient.  This comparison is performed using the inverse approximate Hessian as the metric.  That is, we increment the active subset whenever
\begin{align}
\left(\bar{f}'^t\right)^T {\mb H^t}^{-1} {\bar{f}'^t} < \alpha \frac{
	\sum_i  \left(f'^t_i\right)^T {\mb H^t}^{-1} f'^t_i
	}{
	\left(N^t-1\right)N^t
	}
,
\end{align}
where $N^t$ is the size of the active subset at time $t$, $\mb H^t$ is the full Hessian, and $\bar{f}'^t$ is the average gradient,
\begin{align}
\bar{f}'^t &= \frac{1}{N^t} \sum_i {f_i}'\left( \mb x_i^t \right) 
.
\end{align}
For all the experiments shown here, $\alpha = 1$, and the initial active subset size is two.  We additionally increased the active active subset size by 1 when a bad update is detected (Section \ref{sec bad up}) or when a full pass through the active batch occurs without a batch size increase.  See Supplemental Figure \ref{fig design choices} for a comparison to the case where all subfunctions are initially active.

\subsection{Detecting Bad Updates}
\label{app bad up}

For some ill-conditioned problems, such as ICA with a Student's t-prior (see Section \ref{sec exp results}), we additionally found it necessary to identify bad proposed parameter updates.  In BFGS and LBFGS, bad update detection is also performed, but it is achieved via a line search on the full objective.  Since we only evaluate a single subfunction per update step, a line search on the full objective is impossible.  Updates are instead labeled bad when the value of a subfunction has increased since its previous evaluation, and also exceeds its approximating function by more than the corresponding reduction in the summed approximating function (ie $f_j\left( \mb x^t \right) - g_j^{t-1}\left( \mb x^t \right) > G^{t-1}\left( \mb x^{t-1} \right) - G^{t-1}\left( \mb x^t \right)$).
% We emphasize that the modifications in this section are only required for problems that are ill-conditioned and non-convex.  

When a bad update proposal is detected, $\mb x^t$ is reset to its previous value $\mb x^{t-1}$.  The BFGS history matrices $\Delta {f'}$ and $\Delta \mb x$ are also updated to include the change in gradient in the failed update direction.  Additionally, after a failed update, the update step length in Equation \ref{eq newton step} is temporarily shortened.  It then decays back towards 1 with a time constant of one data pass.  That is, the step length is updated as,
\begin{align}
\label{eq eta update}
\eta^{t+1}
 =
	\left\{\begin{array}{lcrl}
\frac{1}{N} + \frac{N-1}{N} \eta^t & & \text{successful update}  \\
\frac{1}{2} \eta^t & & \text{failed update}
	\end{array}\right.
.
\end{align}
This temporary shortening of the update step length is motivated by the observation that when the approximate Hessian for a single subfunction becomes inaccurate it has often become inaccurate for the remaining subfunctions as well, and failed update steps thus tend to co-occur.

%\newpage
\setcounter{figure}{0} \renewcommand{\thefigure}{C.\arabic{figure}}
\begin{figure*}[htp]
\centering
%\begin{center}
\includegraphics[width=0.3\linewidth]{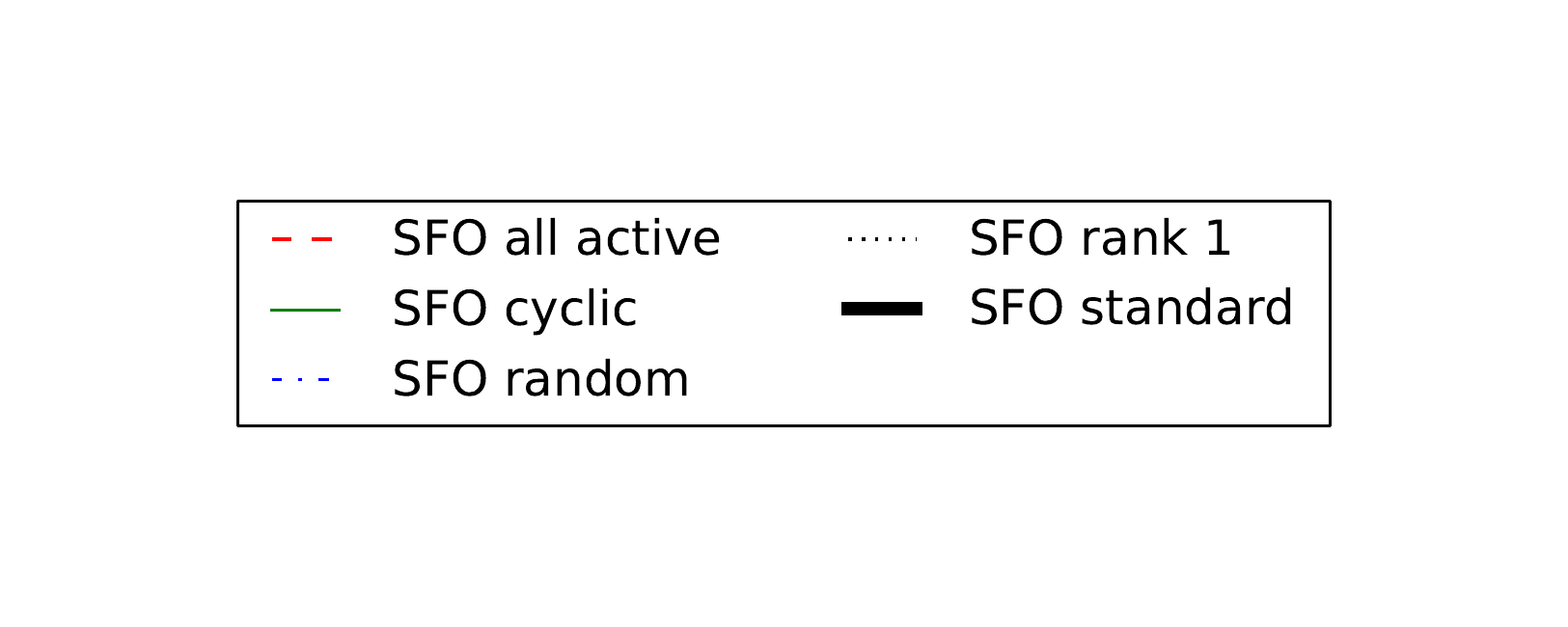} \\
\begin{tabular}{cc}
\hspace{-5mm}
\begin{tabular}{c}
(a)\includegraphics[width=0.3\linewidth]{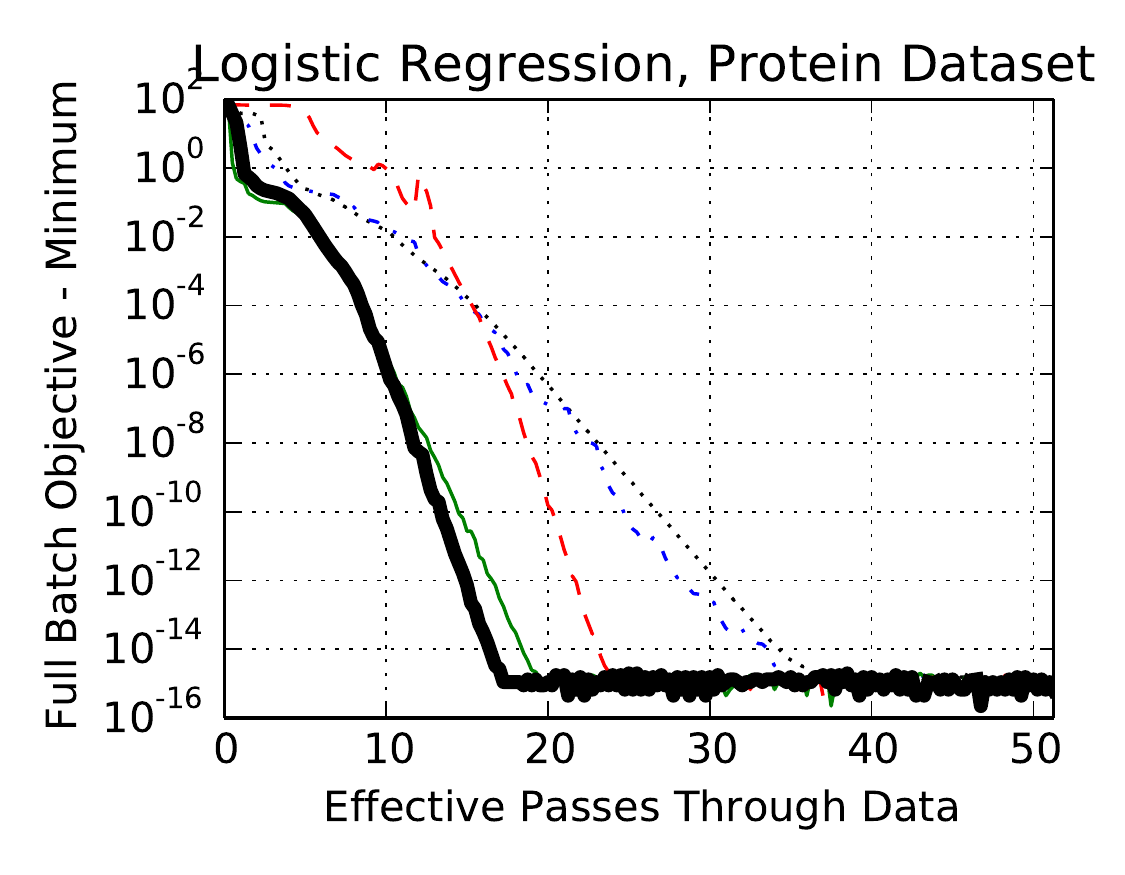}
\end{tabular}
\hspace{-5mm}
 & 
\begin{tabular}{c}
(b)\includegraphics[width=0.3\linewidth]{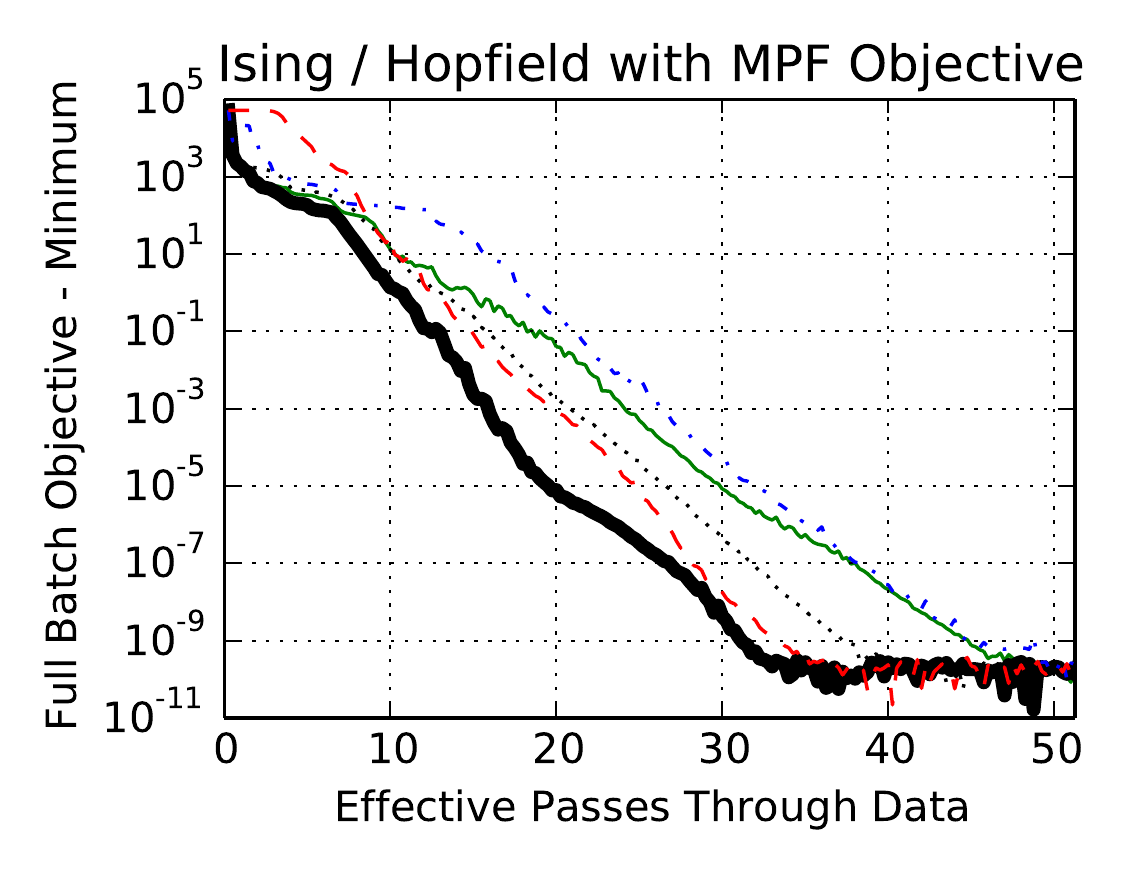}
\end{tabular}
\hspace{-5mm}
\end{tabular}
%\end{center}
\caption{
SFO is insensitive to several specific design decisions.  Plots showing the consequences of changing several of the design choices made in SFO for {\em (a)} the logistic regression objective, and {\em (b)} the Ising / Hopfield objective.  SFO as described in this paper corresponds to the {\em SFO standard} line.  All other lines correspond to changing a single design choice.  {\em SFO rank 1} corresponds to using a rank 1 rather than BFGS update (Section \ref{sec online hessian}).  {\em SFO all active} corresponds to  starting optimization with all the subfunctions active (Section \ref{sec active growth}).  {\em SFO random} and {\em SFO cyclic} correspond to random and cyclic update ordering, rather than maximum distance ordering (Section \ref{sec order}).  For all design choices, SFO outperforms all other techniques in Figure \ref{fig results}.
}
\label{fig design choices}
\end{figure*}

\setcounter{table}{0} \renewcommand{\thetable}{A.\arabic{table}}
\begin{table*}
\begin{center}\begin{tabular}{llll}
\em Operation & \em One time cost & \em Repeats per pass & \em Cost per pass \\
\hline \\ \vspace{-6mm} \\
Function and gradient computation & $\mc O\left( Q \right)$  & $\mc O\left( N \right)$  & $\mc O\left( QN \right)$ \\
Subspace projection & $\mc O\left( M N \right)$  & $\mc O\left( N \right)$  & $\mc O\left( M N^2 \right)$ \\
Subspace collapse & $\mc O\left( M N^{1.4} + N^3 \right)$  & $\mc O\left( 1 \right)$  & $\mc O\left( M N^{1.4} + N^3 \right)$ \\
Minimize $G^t\left(\mb x\right)$ & $\leq \mc O\left( N^{2.4} \right)$  & $\mc O\left( N \right)$  & $\leq \mc O\left( N^{3.4} \right)$ \\
BFGS & $\le \mc O\left( N L^{2} + L^{3}\right)$  & $\mc O\left( N \right)$  & $\le \mc O\left( N^2 L^{2} + N L^{3}\right)$ \\
%BFGS $B_s$, all $s>0$ & $\mc O\left( N L \right)$  & $\mc O\left( N \right)$  & $\mc O\left( N^2 L \right)$ \\
%Natural gradient update & $\mc O\left( M N \right)$  & $\mc O\left( N \right)$  & $\mc O\left( M N^2 \right)$ \\
%Natural gradient subspace update & $\mc O\left( M N^2 \right)$  & $\mc O\left( 1 \right)$  & $\mc O\left( M N^2 \right)$ \\
\hline \\ \vspace{-7mm} \\
Total &    &    & $\mc O\left( QN + M N^2 + N^{3.4} + N^2 L^2 + N L^3 \right)$ \\
\end{tabular} \caption{Computational cost for components of SFO.  $Q$ is the cost of evaluating the objective function and gradient for a single subfunction, $M$ is the number of parameter dimensions, $N$ is the number of subfunctions, $L$ is the number of history terms kept per subfunction.  Typically, $M \gg N \gg L$.  In this case all contributions besides $\mc O\left( QN + M N^2 \right)$ become small.  Additionally the number of subfunctions can be chosen such that $\mc O\left(QN\right) = \mc O\left( MN^2\right)$ (see Section \ref{sec ideal} in the main paper).  Contributions labeled with `$\le$' could be reduced with small changes in implementation, as described in Section \ref{sec cost appendix}.  However as also discussed, and as illustrated by Figure \ref{fig overhead} in the main text, they are not typically the leading terms in the cost of the algorithm.
%*Minimize G currently implemented less efficiently than this
\label{tb cost}
}
\end{center}
\end{table*}

\end{document}